\documentclass[acmtog]{acmart}
\acmSubmissionID{1207}

\usepackage{algorithm}
\usepackage{algpseudocode}

\algrenewcommand\algorithmiccomment[1]{\hfill$\triangleright$ #1}
\usepackage{multirow}
\usepackage{tikz}
\usepackage{array}
\usepackage{adjustbox}
\usepackage[normalem]{ulem} 

\usepackage{xspace}
\usepackage{xcolor}
\usepackage{float} 
\usepackage{subcaption}    
\usepackage{booktabs}

\newcommand{\ourmodel}{PartUV\xspace}

\author{Zhaoning Wang}
\affiliation{
  \institution{Hillbot Inc.}
  \country{USA}
}
\email{zhaoning.eric.wang@gmail.com}

\author{Xinyue Wei}
\affiliation{
  \institution{University of California San Diego}
  \country{USA}
}
\affiliation{
  \institution{Hillbot Inc.}
  \country{USA}
}
\email{xiwei@ucsd.edu}

\author{Ruoxi Shi}
\affiliation{
  \institution{University of California San Diego}
  \country{USA}
}
\affiliation{
  \institution{Hillbot Inc.}
  \country{USA}
}
\email{r8shi@ucsd.edu}

\author{Xiaoshuai Zhang}
\affiliation{
  \institution{Hillbot Inc.}
  \country{USA}
}
\email{x@hillbot.ai}

\author{Hao Su}
\affiliation{
  \institution{University of California San Diego}
  \country{USA}
}
\affiliation{
  \institution{Hillbot Inc.}
  \country{USA}
}
\email{haosu@ucsd.edu}

\author{Minghua Liu}
\affiliation{
  \institution{Hillbot Inc.}
  \country{USA}
}
\email{m@hillbot.ai}

\copyrightyear{2025}
\acmYear{2025}
\setcopyright{acmlicensed}\acmConference[SA Conference Papers '25]{SIGGRAPH Asia 2025 Conference Papers}{December 15--18, 2025}{Hong Kong, Hong Kong}
\acmBooktitle{SIGGRAPH Asia 2025 Conference Papers (SA Conference Papers '25), December 15--18, 2025, Hong Kong, Hong Kong}
\acmDOI{10.1145/3757377.3763843}
\acmISBN{979-8-4007-2137-3/2025/12}


\begin{document}

\title{PartUV: Part-Based UV Unwrapping of 3D Meshes}

\begin{teaserfigure}
  \includegraphics[width=\textwidth]{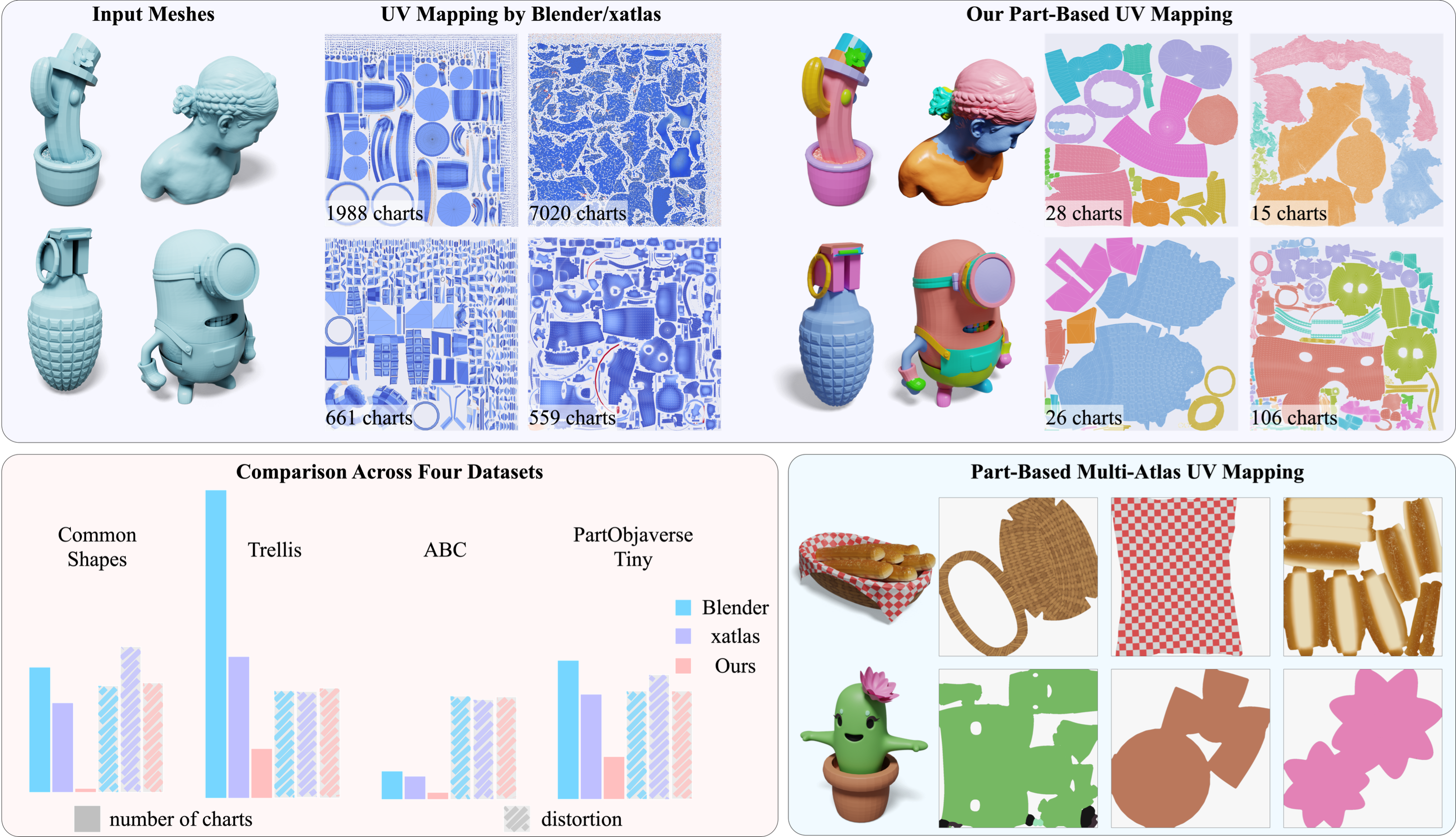}
  \vspace{-2.5em}
  \caption{We propose PartUV, a novel part-based UV unwrapping method for 3D meshes. Unlike traditional approaches that rely solely on local geometric priors and often produce over-fragmented charts, PartUV combines learned part priors with geometric cues to generate a small number of part-aligned charts. We evaluate our method on four diverse datasets—PartObjaverseTiny (man-made)~\cite{yang2024sampart3d}, Trellis (AI-generated)~\cite{xiang2024structured}, ABC (CAD)~\cite{koch2019abc}, and Common Shapes~\cite{jacobson_common3d_2023}—and show that it produces significantly less fragmented UV mappings while maintaining low distortion on par with baseline methods. Leveraging part-aware charts also enables applications such as generating one atlas per part.
  }
  \label{fig:teaser}
\end{teaserfigure}

\begin{abstract}
UV unwrapping flattens 3D surfaces to 2D with minimal distortion, often requiring the complex surface to be decomposed into multiple charts. Although extensively studied, existing UV unwrapping methods frequently struggle with AI-generated meshes, which are typically noisy, bumpy, and poorly conditioned. These methods often produce highly fragmented charts and suboptimal boundaries, introducing artifacts and hindering downstream tasks. We introduce \ourmodel, a part-based UV unwrapping pipeline that generates significantly fewer, part-aligned charts while maintaining low distortion. Built on top of a recent learning-based part decomposition method PartField, \ourmodel combines high-level semantic part decomposition with novel geometric heuristics in a top-down recursive framework. It ensures each chart’s distortion remains below a user-specified threshold while minimizing the total number of charts. The pipeline integrates and extends parameterization and packing algorithms, incorporates dedicated handling of non-manifold and degenerate meshes, and is extensively parallelized for efficiency. Evaluated across four diverse datasets—including man-made, CAD, AI-generated, and Common Shapes—\ourmodel outperforms existing tools and recent neural methods in chart count and seam length, achieves comparable distortion, exhibits high success rates on challenging meshes, and enables new applications like part-specific multi-tiles packing. Our project page is at \url{https://www.zhaoningwang.com/PartUV}.

\end{abstract}

\maketitle

\section{Introduction}
UV unwrapping projects the 3D surface of a mesh onto a 2D plane, assigning every 3D surface point a corresponding 2D UV coordinate. This step is fundamental in 3D‑content‑creation pipelines because it enables detailed surface information-such as material properties (e.g., base‑color, roughness, normal maps), along with auxiliary maps like ambient occlusion and displacement—to be efficiently stored and manipulated in 2D space.

A principal component of UV unwrapping is surface parameterization, which flattens the 3D surface while trying to preserve geometric properties such as angles and areas. For meshes with complex geometry, flattening the entire surface into a single 2D chart introduces large distortion. Consequently, chart segmentation (or seam cutting) is typically used to divide the mesh into multiple charts along strategically placed seams, allowing each chart to be flattened with reduced distortion. Finally, UV packing arranges the resulting charts within the unit square (the UV atlas) to maximize texture‑space use.

Although UV unwrapping has been studied extensively, existing methods are typically tuned for well-behaved meshes, such as those created by professional 3D artists. They often fail on more complex, AI‑generated meshes. Such meshes, typically extracted from neural‑field isosurfaces (e.g., via marching cubes~\cite{lorensen1998marching}), tend to have bumpy surfaces, many small triangles, and poor geometric quality (e.g., disconnected components or holes). On such data, existing methods may time‑out or return extremely fragmented atlases in which a single chart holds only one or a handful of triangles. This extreme fragmentation hampers texture painting and editing, introduces texture‑bleed and baking or rendering artifacts at chart boundaries, and burdens downstream applications with an unwieldy number of charts.

Some recent approaches mitigate fragmentation~\cite{srinivasan2024nuvo,zhang2024flatten,zhao2025flexpara} by representing the UV mapping using a neural field and optimizing such a field for each 3D shape from scratch. While these methods can effectively control the number of charts generated, they typically run for more than thirty minutes and still exhibit noticeable distortion. Other methods~\cite{li2018optcuts, autoCuts} jointly optimize seam length and distortion but are likewise computationally expensive. Moreover, some existing approaches like~\cite{sorkine2002bounded,lscm, zhou2004iso} segment charts or identify seams using heuristics based on local geometric properties, rather than leveraging the concept of geometric or semantic parts. This can lead to unintuitive or suboptimal chart boundaries that split semantically coherent regions across multiple charts, further complicating downstream tasks such as texture authoring, part-based editing, and semantic-aware rendering.

In this paper, we introduce \ourmodel, a part-based UV unwrapping pipeline for 3D meshes that generates UV mappings with a small number of part-aligned charts while maintaining low distortion, as well as robust and efficient processing—typically completed within a few to several tens of seconds. \ourmodel builds on a recent learning‑based method, PartField~\cite{liu2025partfield}, which produces a hierarchical part tree for the input mesh. \ourmodel also proposes several novel geometric heuristics that further decompose simple local parts into charts that can be flattened with minimal distortion. Combining high‑level semantic decomposition from PartField with fine‑grained geometric cuts, \ourmodel employs a top‑down recursive search that minimizes chart count while keeping each chart’s distortion below a user‑specified threshold. \ourmodel uses established surface parameterization algorithms (e.g., ABF++~\cite{sheffer2005abf++}) for chart flattening and proven packing algorithms for optimal atlas layout. To ensure high speed and robustness, the pipeline incorporates extensive parallelization and acceleration strategies, as well as dedicated handling of non-manifold and degenerate cases.

By explicitly incorporating semantic priors, our approach yields several key benefits. First, semantics improve decomposition by reducing excessive reliance on local geometry, which often causes over-segmentation and long runtimes. Second, semantic cues preserve the coherence of object parts, preventing chart boundaries from cutting through meaningful regions (e.g., across the flat surface of a TV screen, as in Figure~\ref{fig:partuv_logo}, or a human face), thereby facilitating editing and rendering tasks. Third, semantic grouping naturally supports better chart packing strategies, allowing related charts to be organized together—either within a single atlas (top right of Figure~\ref{fig:teaser}) or across multiple atlases (bottom right of Figure~\ref{fig:teaser}). This enhances organizational clarity and simplifies the process of locating and editing related charts. Finally, seams guided by semantic boundaries tend to fall in perceptually unobtrusive locations, improving the visual quality of textured models.

We evaluate \ourmodel on four datasets spanning man-made, AI-generated, CAD models, and common 3D shapes (e.g., Stanford Bunny, XYZ Dragon), and compare it against widely used UV unwrapping tools such as xatlas~\cite{xatlas}, Blender~\cite{blender}, and Open3D~\cite{zhou2018open3d}, as well as the recent neural-based methods~\cite{srinivasan2024nuvo}. As shown in Figure~\ref{fig:teaser}, our method decomposes input meshes into significantly fewer charts—also resulting in shorter seam lengths—while maintaining low angular and area distortion comparable to baseline methods. The incorporation of explicit part priors not only helps the segmented charts better align with part boundaries but also enables new applications, such as packing semantic-aligned parts into separate atlases. Moreover, \ourmodel maintains a high success rate, handles a wide range of meshes (including large, complex, and non-manifold ones), and processes each mesh quickly—typically within a few to several tens of seconds.

\begin{figure*}
    \centering
    \includegraphics[width=\linewidth]{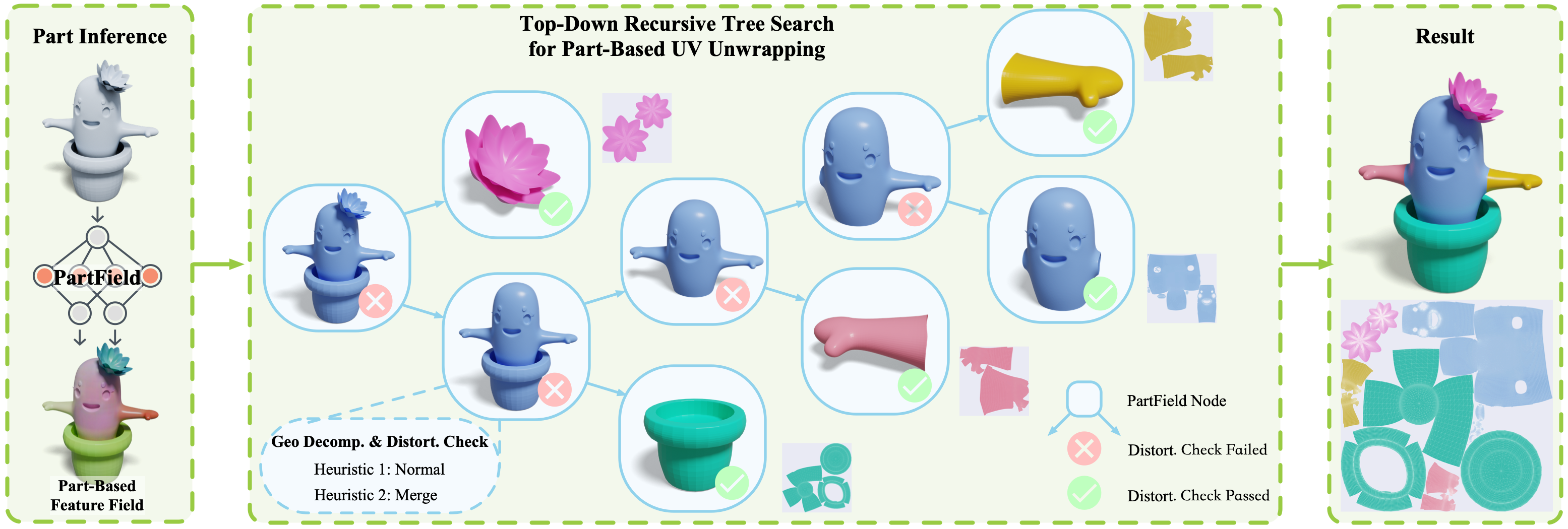}
    \vspace{-2.5em}
    \caption{\textbf{Pipeline of PartUV.} Given a mesh $\mathcal{M}$, we first leverage the learning-based method PartField to predict a part-aware feature field. By applying a clustering algorithm to this field, we obtain a hierarchical part tree $\mathcal{T}$ for the input mesh $\mathcal{M}$. We then recursively traverse the part tree $\mathcal{T}$ starting from the root node. For each visited node $\mathcal{P}$, we apply two novel geometry-based strategies to segment the corresponding part mesh into multiple sets of charts $\mathscr{C}$. Each chart in the set is then flattened using the ABF algorithm~\cite{sheffer2005abf++}, and its distortion is evaluated. If the distortion exceeds a user-specified threshold $\tau$, we recursively traverse the left and right children of the part tree $\mathcal{T}$. Otherwise, we adopt the segmented charts and their corresponding UV mappings for the part mesh.}
    \vspace{-3.5ex}
    \label{fig:pipeline}
\end{figure*}
\section{Related Work}
\subsection{Mesh Parameterization}

Disregarding mesh decomposition, mesh parameterization—the process of mapping a 3D mesh to a 2D domain while minimizing various distortion metrics (e.g., isometric, conformal, equiareal)—is a fundamental operation in UV unwrapping~\cite{sawhney2017boundary,schuller2013locally,rabinovich2017scalable}. Many methods directly optimize UV coordinates based on mesh connectivity by solving linear or nonlinear systems under fixed or free boundary conditions. These include barycentric embeddings like Tutte’s embedding~\cite{tutte1963draw}, Laplacian eigen-decomposition techniques~\cite{belkin2003laplacian,mullen2008spectral,taubin1995signal}, and distortion-minimization methods~\cite{hormann2000mips,sander2001texture,su2016area,yueh2019novel,yan2005mesh,desbrun2002intrinsic,Ray2003HierarchicalLS,ben2008conformal}, such as Least Squares Conformal Maps (LSCM)~\cite{lscm}. Some adopt local/global strategies that combine per-triangle transformations (e.g., rotations) with global stitching~\cite{sorkine2007rigid,liu2008local,alexa2023rigid}. In contrast, Angle-Based Flattening (ABF) methods~\cite{sheffer2001parameterization,sheffer2000surface,sheffer2005abf++,zayer2007linear} optimize triangle angles before converting them into UVs, typically offering better angle preservation than LSCM but at higher computational cost due to nonlinear solvers. In this paper, we adopt ABF for base unfolding, integrating it with various mesh decomposition strategies.

\subsection{Mesh Segmentation for UV Unwrapping}
Mesh segmentation is a key step in UV unwrapping, aiming to divide a complex 3D mesh into simpler patches, or charts, each of which can be flattened with minimal distortion. Existing approaches can be broadly categorized into four main strategies. Top-down methods recursively subdivide the mesh, often using spectral analysis~\cite{liu2007mesh,pothen1990partitioning,Liu2007MeshSegmentation} or cuts along feature lines~\cite{zhang2005feature} and high-curvature regions~\cite{lavoue2005new}. Bottom-up aggregation~\cite{sorkine2002bounded,lscm,kalvin1996superfaces,julius2005d, pulla2001improved, zhou2004iso, yamauchi2005mesh,takahashi2011optimized,bhargava2025mesh} grows charts from small seeds (e.g., triangles), merging them based on planarity or distortion bounds to maximize chart size under quality constraints. Cut optimization~\cite{li2018optcuts, autoCuts,pietroni2009almost,gu2002geometry,Sander2003MultiChartGI, takahashi2011optimized, carr2006rectangular} focuses on identifying optimal seam networks that minimize seam length, reduce flattening distortion, or better suit specific parameterization goals. Clustering-based methods~\cite{katz2003hierarchical,roy2023neural} group mesh elements based on geometric similarity to form coherent segments. Commonly used tools such as xatlas~\cite{xatlas}, Blender~\cite{blender}, and Open3D~\cite{zhou2018open3d} primarily adopt bottom-up aggregation strategies that rely solely on local geometric properties. In contrast, PartUV integrates high-level semantic part priors from a learning-based module with novel local geometric decomposition strategies via a top-down recursive tree search.

\subsection{Learning-Based Parameterization and Segmentation}

Representing UV mappings with neural networks has attracted attention due to their differentiability and composability. Neural Surface Maps~\cite{morreale2021neural} pioneered this direction by learning mappings across collections of surfaces. Several approaches jointly optimize UV parameterization and 3D reconstruction from multi-view images using cycle-consistency and distortion-based losses~\cite{xiang2021neutex,LearninganIsometric}. Others, such as NUVO~\cite{srinivasan2024nuvo}, solve for parameterizations directly from sampled 3D points. More recent methods~\cite{zhang2024flatten,zhao2025flexpara} design neural networks to emulate physical operations, including face cutting, UV deformation, and unwrapping. However, all the aforementioned techniques typically rely on per-shape optimization, making them computationally expensive and often requiring tens of minutes per shape. A few feedforward models address specific tasks more efficiently, such as intra-category texture transfer~\cite{chen2022auv} and low-distortion patch selection~\cite{liu2023wand}.

To enhance UV chart decomposition, we explore the integration of semantic part priors. Traditional learning-based part segmentation methods~\cite{qian2022pointnext,jiang2020pointgroup,vu2022softgroup} are restricted to closed-set categories due to the limited scale of part-annotated datasets~\cite{mo2019partnet,yi2016scalable}. Recent approaches~\cite{yang2023sam3d,xu2023sampro3d,yin2024sai3d,guo2024sam,zhou2018open3d,xu2024embodiedsam,he2024pointseg} lifting priors from 2D vision to 3D models~\cite{kirillov2023segment,li2022grounded} for open-world 3D part segmentation but rely on per-shape optimization, leading to slow runtimes and noise sensitivity. In contrast, a recent method, PartField~\cite{liu2025partfield}, introduces a feedforward model that learns part-aware feature fields for fast, hierarchical 3D part decomposition. While semantic priors intuitively benefit UV unwrapping, naïve integration can yield suboptimal results. PartUV addresses this with a novel top-down strategy that interleaves semantic segmentation and geometric flattening.
\section{Method}
\subsection{Overview}
For a 3D mesh $\mathcal{M} = (V,F)$, \ourmodel decomposes the mesh faces into a small collection of disjoint charts:
\vspace{-0.5em}
\begin{equation}
F = \bigcup_{k=1}^{K} C_k, \quad \text{with} \quad C_i \cap C_j = \varnothing \quad (i \neq j),
\vspace{-1ex}
\end{equation}
where each \textbf{\emph{chart}} $C_k$ is a connected subset of faces. We utilize an Angle-Based Flattening algorithm(ABF++), to flatten each chart onto a 2D plane, yielding mappings:
\vspace{-0.5em}
\begin{equation}
\phi_k : C_k \longrightarrow \mathbb{R}^2,
\vspace{-1ex}
\end{equation}
so that each vertex $v_i \in V$ receives a corresponding 2D UV coordinate $\mathbf{u}_i = \phi_k(v_i) \in \mathbb{R}^2$, determined by the chart it belongs.

The primary challenge lies in decomposing the mesh into charts $\{C_k\}$, a task essentially equivalent to identifying optimal seams for mesh segmentation. Many state-of-the-art methods, such as xatlas and Blender's Smart UV, rely on geometric heuristics—such as bottom-up greedy chart expansion based on face normals or distortion metrics—to generate these charts. However, these strategies typically produce overly fragmented charts and unintuitive or suboptimal boundaries, often splitting semantically coherent regions across multiple charts.

In contrast, \ourmodel adopts a coarse-to-fine, two-stage strategy for mesh decomposition. At a high level, it employs a recent learning-based method called PartField (introduced in Section~\ref{sec:partfield}) to partition mesh faces into semantically coherent \textbf{\emph{parts}}: $F = \bigcup_{m=1}^{M} P_m$, where each part $P_k$ exhibits relatively simple geometry—for instance, cylindrical limbs, spherical toes. Subsequently, we introduce two geometry-based heuristics (detailed in Section~\ref{sec:heuristic}) to further segment each part into a small set of \textbf{\emph{charts}}:
$P_m = \bigcup_{n=1}^{N_m} C_{m,n}$, ensuring each resulting chart $C_{m,n}$ can be flattened onto a 2D plane with minimal distortion.

To accomplish this, \ourmodel employs a top-down recursive decomposition search (detailed in Section~\ref{sec:search}) that minimizes the total number of generated charts while ensuring the distortion for each chart remains below a user-specified threshold $\tau$. Since exhaustive decomposition searches can incur substantial computational overhead, we introduce acceleration and parallelization techniques for efficiency. To ensure a complete and robust pipeline, Section~\ref{sec:runtime} also describes additional preprocessing and postprocessing procedures, including handling non-manifold and multi-component meshes, as well as performing UV packing.

\subsection{Preliminary: PartField}
\label{sec:partfield}

PartField~\cite{liu2025partfield} trains a feed‑forward neural network that takes a 3D mesh as input and predicts a continuous, part-based feature field encoded as a triplane. By leveraging extensive contrastive learning on part-labeled 3D data and large‑scale unlabeled 3D data with 2D pseudo part labels, PartField learns general hierarchical concepts of semantic and geometric parts. For any 3D point, we can obtain its high‑dimensional part feature by interpolating the triplane representation. Points whose features are similar—as measured by cosine similarity—are therefore more likely to belong to the same part.
\subsection{Top-Down Recursive Tree Search}
\label{sec:search}

After obtaining the hierarchical part-based feature field using PartField, we first compute a representative part feature for each triangular face by uniformly sampling $s$ 3D points within the face and averaging their corresponding point features. Next, we construct a hierarchical part tree $\mathcal{T}$ using agglomerative clustering~\cite{johnson1967hierarchical} on these face features. In this hierarchical tree structure, the leaf nodes represent individual triangular faces, and the root node encompasses the entire mesh.

% ------------------------------------------------------------------
\begin{algorithm}[t]
\caption{\textsc{PartTreeSearch}}
\label{alg:search}
\begin{algorithmic}[1]
\Require $\mathcal{P}$ (a PartField subtree node), distortion threshold $\tau$, chart budget $B$
\Ensure A \emph{UVChartSet} $\mathscr{C}=\{C_1, C_2, \cdots, C_k\}$ for $\mathcal{P}$'s mesh, whose charts $C_i$ have distortion $\le \tau$ and whose total chart count $k$ is $\le B$; return $\bot$ if no such set exists.

\vspace{2pt}

\Procedure{PartTreeSearch}{$\mathcal{P}$,\,$\tau$,\,$B$}
    \If{$B < 1$}                          \Comment{budget exhausted}
        \State \Return $\bot$
    \EndIf
% ------------------------------------------------------------------
    \State $\mathcal{H}_1 \gets \Call{GenCandidatesH1}{\mathcal{P}.mesh,10}$ \Comment{Heuristic1} \label{line:heuristic1}
    \ForAll{$\mathscr{C} \in \mathcal{H}_1$} \Comment{$\mathscr{C}$ is one of candidate $UVChartSet$}
        \State \Call{ParameterizeABF}{$\mathscr{C}$}   \Comment{flatten and compute distortion}
    \EndFor
% ------------------------------------------------------------------
    \If{$\min_{\mathscr{C}\in \mathcal{H}_1} \mathscr{C}.\text{dist} > \tau$}            \Comment{no admissible H1 candidate}
        \State $L \gets \Call{PartTreeSearch}{\mathcal{P}.left\_child ,\tau, \infty\,}$
        \State $R \gets \Call{PartTreeSearch}{\mathcal{P}.right\_child ,\tau, \infty\,}$
        \State \Return $L \oplus R$ \Comment{merge chart sets from the subtrees}
    \Else
% ------------------------------------------------------------------
    \State $\mathcal{H}_2 \gets \{\Call{GenCandidateH2}{\mathcal{P}.mesh,\tau}\}$ \Comment{Heuristic2} \label{line:heuristic2}
    \State \Call{ParameterizeABF}{$\mathcal{H}_2[0]$} 
    \State $\mathcal{S} \gets \bigl\{\,\mathscr{C}\in \mathcal{H}_1 \cup \mathcal{H}_2 \mid
            \mathscr{C}.\text{dist}\le\tau \;\land\; \Call{NoOverlap}{\mathscr{C}}\bigr\}$
    \State $\mathscr{C}_{\text{best}} \gets \arg\!\min_{\mathscr{C} \in \mathcal{S}} \mathscr{C}.\text{count}$
% ------------------------------------------------------------------
    \State \Comment{Recurse to determine whether a better solution exists}     \label{line:recursion}
    \State $B' \gets \min(B,\,\mathscr{C}_{\text{best}}.\text{count}-1)$ 
    \State $L \gets \Call{PartTreeSearch}{\mathcal{P}.left ,\tau,B'-1}$
    \State $R \gets \Call{PartTreeSearch}{\mathcal{P}.right,\tau,B'-L.\text{count}}$
    \State $\mathscr{C}_{\text{comb}} \gets L \oplus R$
    \State $is\_valid \gets \mathscr{C}_{\text{comb}}.\text{dist}\le\tau \;\land\; \Call{NoOverlap}{\mathscr{C}_{\text{comb}}}$
    \If{$\mathscr{C}_{\text{comb}}.\text{count}<\mathscr{C}_{\text{best}}.\text{count} \;\land\; is\_valid$}
        \State $\mathscr{C}_{\text{best}}\gets \mathscr{C}_{\text{comb}}$
    \EndIf
    \State \Return $\mathscr{C}_{\text{best}}$
    \EndIf
% ------------------------------------------------------------------

\EndProcedure
\end{algorithmic}

\end{algorithm}
% \vspace{-2ex}

A straightforward approach to mesh decomposition might involve using PartField to directly generate a fixed number of parts and attempting to flatten each to 2D individually. Alternatively, one could adaptively traverse $\mathcal{T}$ from the root downward, checking whether each node’s corresponding geometry can be flattened into a 2D chart with minimal distortion. However, we observe that relying solely on parts generated by PartField often leads to suboptimal results. This limitation arises because PartField primarily focuses on semantic or coarse geometric partitioning and is less effective for finer-scale, UV-related decompositions—such as accurately segmenting cylindrical or spherical regions into charts that minimize distortion. To address this challenge, \ourmodel leverages PartField primarily for high-level semantic decomposition into geometrically simpler subparts and employs two geometry-based heuristics (elaborated in Section~\ref{sec:heuristic}) to further divide these semantically meaningful parts into smaller charts amenable to low-distortion flattening. The search algorithm interleaves these two strategies.

Formally, \ourmodel employs a top-down recursive decomposition search—detailed in Algorithm~\ref{alg:search}—to optimally balance the chart count against distortion constraints. Given a subtree node $\mathcal{P}$ (initially the root node of the tree $\mathcal{T}$), a distortion threshold $\tau$, and a chart budget $B$ (initially set to $\infty$), the algorithm searches for a valid chart decomposition of $\mathcal{P}$'s mesh that satisfies three key conditions: each chart has a distortion of at most $\tau$, no charts overlap in 2D, and the total number of charts does not exceed $B$. If no feasible decomposition exists, the algorithm returns failure ($\bot$).

Specifically, the algorithm begins by generating candidate chart decompositions for node $\mathcal{P}$'s mesh using a primary geometric heuristic (Heuristic1). Each candidate chart set $\mathscr{C} = \{C_i\}$, consisting of up to $t$ charts, is flattened using Angle-Based Flattening (ABF), and its distortion is evaluated. If none of the candidate decompositions with up to $t$ charts from Heuristic1 satisfy the distortion constraint (i.e., $\delta_{\min} > \tau$), the algorithm utilizes the PartField tree to divide the mesh and recursively searches both the left and right child subtrees without budget constraints, then merges their respective optimal chart sets.

Conversely, if an admissible candidate decomposition is found through Heuristic1, we further refine it using a secondary heuristic (Heuristic2) designed to potentially yield fewer charts. Among these candidates, we select the best solution $\mathscr{C}_{\text{best}}$ with the minimal chart count that satisfies the distortion and overlap constraints.

Before accepting this solution, the algorithm performs a final check by recursively exploring the left and right child nodes of the PartField subtree, using a reduced chart budget ($B'$), which is derived from the best candidate’s chart count minus one. If the combined solution from this subtree search yields a valid and superior decomposition (i.e., fewer charts than $\mathscr{C}_{\text{best}}$), it replaces the previously identified best solution. The chart budget $B$ prevents the search from going too deep. As the search progresses, $B$ tightens, and recursion stops when candidates exceed it. In practice, the search rarely goes very deep.

By strategically interleaving hierarchical semantic guidance from PartField with fine-grained geometric heuristics and systematic recursive exploration, our proposed decomposition search achieves semantically coherent, distortion-bounded, and notably compact mesh parameterizations—characterized by a small number of charts.

\subsection{Geometry-Based Part Decomposition}
\label{sec:heuristic}
In this section, we introduce two heuristics, \emph{Normal} and \emph{Merge}, to further decompose a part $P_m$ generated by PartField and exhibiting simple geometry. The goal is to divide $P_m$ into multiple charts, denoted as $P_m = \bigcup_{n=1}^{N_m} C_{m,n}$, such that each chart can be flattened to 2D with low distortion.

The first heuristic, referred to as \emph{Normal} (line~\ref{line:heuristic1} in Algorithm~\ref{alg:search}), is based on face normals. We apply an agglomerative clustering algorithm~\cite{johnson1967hierarchical} to the face normals of $P_m$'s mesh, partitioning its triangle faces into charts, where each chart is composed of connected faces with similar normals. This clustering is performed once for $P_m$, producing $t$ candidate decompositions with 1 to $t$ charts (we use $t=10$ in our experiments). For each candidate decomposition, we apply the Angle-Based Flattening (ABF) algorithm to flatten the charts and evaluate distortion. Since ABF aims to preserve angles, we quantify distortion using an area stretch metric, defined as:
\vspace{-0.3em}
\begin{equation}
\label{equ:distortion}
\small
\text{distortion}(\mathscr{C}) = \max_{C \in \mathscr{C}} \left( \frac{1}{|C|} \sum_{f \in C} \max\left( \text{stretch}(f), \frac{1}{\text{stretch}(f)} \right) \right),
\end{equation}
\vspace{-0.3em}
where $C$ denotes a single chart composed of multiple connected faces, and $\mathscr{C}$ denotes the set of all charts in the decomposition. The stretch of a triangle face $f$ is computed as:
\vspace{-0.4em}
% \begin{equation}
% \small
% \text{stretch}(f) = \frac{\text{area}{2D}(f)}{\text{area}{3D}(f)} \Bigg/ \left( \frac{1}{|C|} \sum_{f' \in C} \frac{\text{area}{2D}(f')}{\text{area}{3D}(f')} \right).
% \end{equation}
\begin{equation}
\small
\text{stretch}(f) = \frac{\text{area}{2D}(f)}{\text{area}{3D}(f)} \Bigg/ \left( \frac{\sum_{f' \in C} \text{area}{2D}(f')}{\sum_{f' \in C} \text{area}{3D}(f')} \right).
\end{equation}
Note that both PartField and the \emph{Normal} heuristic employ the agglomerative clustering algorithm to group faces into parts or charts. The key difference lies in the features used: PartField utilizes learned high-level part features, while heuristic \emph{Normal} relies on low-level geometric face normals. These two strategies are thus consistent in spirit and complementary in practice.

The \emph{Normal} heuristic is simple, fast, and often yields satisfactory results. However, we also propose a second, more computationally expensive heuristic, called \emph{Merge} (line~\ref{line:heuristic2} in Algorithm~\ref{alg:search}), which may produce decompositions with fewer charts. Given a part $P_m$, the \emph{Merge} heuristic begins by computing its oriented bounding box (OBB). It then assigns each triangle face a label from 1 to 6, corresponding to the OBB face normal with which the triangle’s normal is most closely aligned. Using both face connectivity and these labels, we segment the faces into multiple connected components. These components are then sorted by size (small to large), and we iteratively attempt to merge each component with its adjacent neighbors, starting from the one with the longest shared edges. For each merge attempt, we temporarily merge two components and apply ABF to flatten the combined chart. The merge is accepted if the resulting chart satisfies the distortion threshold and is free of overlaps. This merging process continues until no further valid merges can be made, after which the final chart set is returned. Since the \emph{Merge} heuristic often begins with many small components and performs multiple ABF calls during its iterative merging process, it is significantly more expensive than the \emph{Normal} heuristic. However, it may yield better decompositions with fewer charts. Therefore, we only invoke the \emph{Merge} heuristic when the \emph{Normal} heuristic returns a valid (admissible) decomposition.

\subsection{Runtime Optimization and Pre- and Post-processing}
\label{sec:runtime}

\textbf{Runtime Optimization} To ensure efficiency despite the computational cost of recursive decomposition and repeated ABF invocations, we adopt two key strategies during the decomposition process. First, we parallelize all recursive calls to the left and right subtrees during the top-down search, allowing the algorithm to exploit multi-core processing and significantly accelerate the overall decomposition. Second, to avoid the high cost of repeatedly invoking Angle-Based Flattening (ABF) on dense meshes, we employ a GPU-accelerated mesh simplification algorithm~\cite{oh2025pamo} to generate low-resolution approximations of candidate charts. During simplification, the chart boundary is kept fixed to preserve the geometric structure relevant to UV mapping. These simplified charts are used to estimate distortion metrics quickly (i.e., surrogate distortion) during intermediate evaluations. Once the final chart set is determined, ABF is applied to the original high-resolution mesh to produce accurate UV coordinates and distortion measurements.

\noindent \textbf{Non-Manifold and Multi-Component Meshes} Since the ABF algorithm assumes each input chart is a manifold and connected surface, additional processing is required when handling non-manifold or multi-component meshes. For non-manifold meshes, we detect all non-manifold edges—i.e., edges shared by more than two faces—and resolve them by duplicating the shared vertices and splitting each such edge into $N-1$ distinct edges, where $N$ is the number of incident faces, thereby converting the structure into a manifold form suitable for flattening. For meshes containing multiple connected components, we initially proceed with the proposed PartField-guided hierarchical decomposition. If a part mesh consists of disconnected components, we skip heuristic-based decomposition at that level and instead recursively explore the left and right subtrees of the PartField hierarchy. However, if PartField fails to further segment the multiple components after reaching a predefined recursion depth, we fall back to applying the geometric heuristics and ABF flattening to each connected component at that level individually, in order to avoid excessively fragmented decomposition.

\noindent \textbf{UV Packing} While our primary focus is on decomposing 3D meshes into charts and generating corresponding low-distortion 2D parameterizations, our method is fully compatible with a variety of existing UV packing algorithms. A distinguishing feature of our approach is that chart decompositions are grouped based on semantically meaningful parts, enabling more structured and application-aware packing strategies. 
For example, charts belonging to the same part can be grouped together during packing. 
Alternatively, part groups can be packed into an arbitrary number of UV atlas squares (e.g., multiple $[0,1]^2$ spaces) with a semantically balanced distribution across atlases.
This semantic hierarchy not only improves organizational clarity but also benefits downstream applications such as texture painting or editing, where charts belonging to the same part remain spatially close and are easier to manipulate collectively.
\section{Experiments}
\subsection{Implementation Details and Evaluation Setup}

\begin{figure*}[t]
    \centering
    \includegraphics[width=\linewidth]{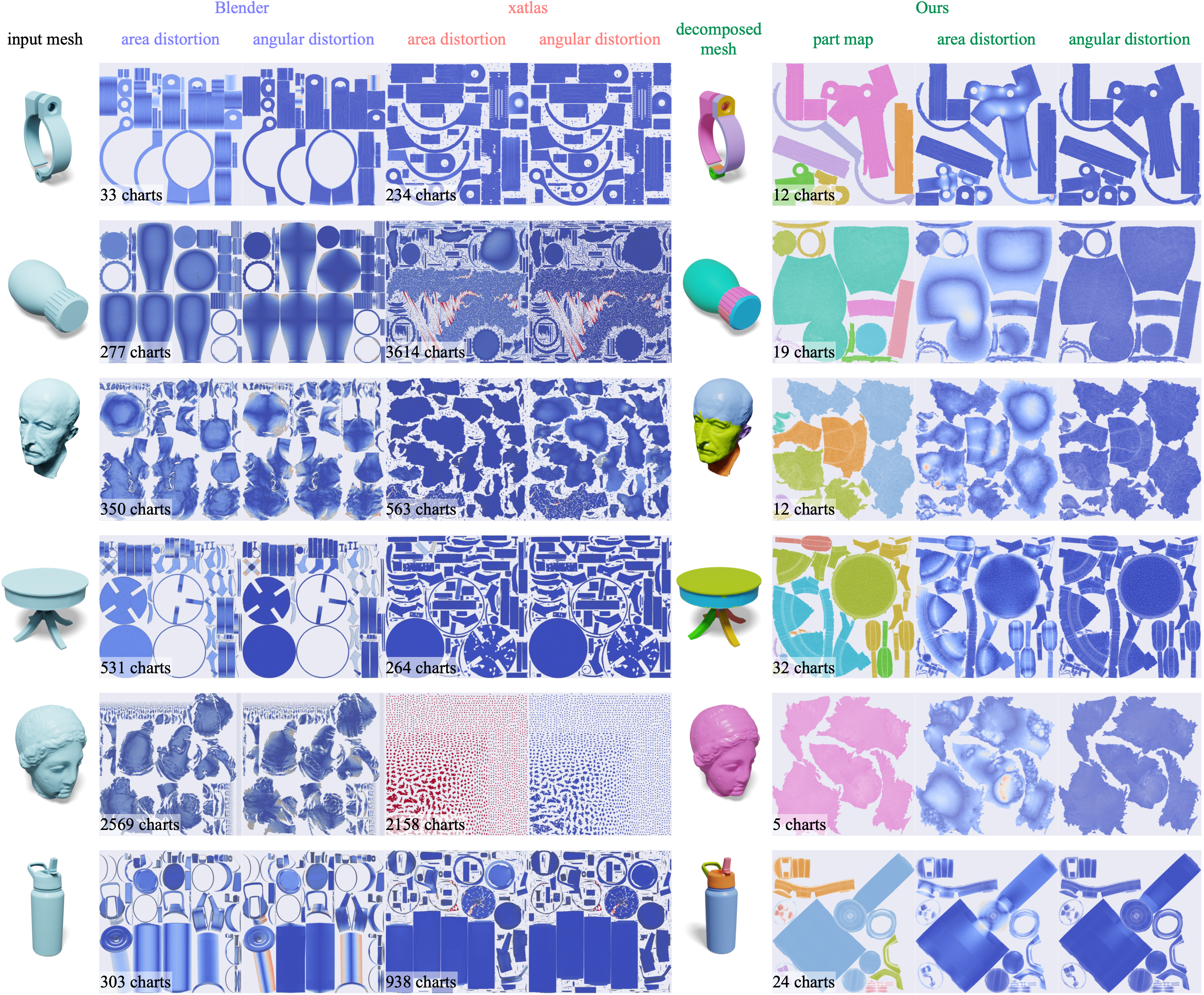} 
    \vspace{-1em}
    \caption{\textbf{Quantitative comparison between Blender~\cite{blender}, xatlas~\cite{xatlas}, and PartUV.} Unlike Blender and xatlas, which rely solely on local geometric properties for UV unwrapping, PartUV integrates high-level semantic priors with low-level geometric heuristics, enabling part-based chart decomposition. As a result, it produces significantly fewer charts, with boundaries that better align with semantic parts.}
    \label{fig:partuv_comparison} 
\end{figure*}

\noindent \textbf{\textbf{Implementation Details.}}
We train PartField~\cite{liu2025partfield} on Objaverse~\cite{objaverse} using 8 NVIDIA H100 GPUs for 15 days. During inference, we sample 10 points per face and average their features for face clustering and part tree construction. The core pipeline of PartUV is implemented in C++17, and we use UVPackmaster~\cite{uvpackmaster3} for final group-based UV packing. A distortion threshold of 1.25 is used in all experiments. For ABF++, we run 5 iterations per call, and follow Blender to set the gradient early-stop condition. Mesh simplification is controlled by a curvature-related threshold of 1e-4 and a maximum iteration count of 1,000. All parameters are fixed across experiments. All methods are evaluated on a cluster node with a 96-core Intel® Xeon® Platinum 8468 CPU and an NVIDIA H100 GPU.

\noindent \textbf{\textbf{Evaluation Datasets.}}
To comprehensively evaluate the approaches across diverse mesh sources, qualities, and styles, we use four datasets: (a) \emph{Common Shapes}~\cite{jacobsonCommon3DTestModels}, a GitHub repository of 24 widely used models in graphics (e.g., Bimba, Igea, Stanford Bunny) with provided processed .obj files;
(b) \emph{PartObjaverseTiny}~\cite{yang2024sampart3d}, a 200-shape subset of Objaverse~\cite{objaverse} featuring high-quality, man-made meshes with multiple components and smooth surfaces; (c) \emph{ABC}~\cite{koch2019abc}, a CAD dataset of mechanical models combining sharp and smooth features—we use the first 100 meshes from its initial batch; and (d) \emph{Trellis}~\cite{xiang2024structured}, which includes 114 AI-generated meshes from a recent 3D diffusion-based generative model. These meshes are typically noisy and geometrically low-quality, posing greater challenges than human-made counterparts.

\noindent \textbf{\textbf{Evaluation Metrics.}}  
We evaluate the quality of the generated UV maps from four perspectives:  
(1) \emph{Number of Charts}: For each shape, we count the number of charts and report both the average and median values across the dataset.  
(2) \emph{Seam Length}: We compute the seam length by summing the lengths of all chart boundary edges, with UV coordinates normalized to a [0, 1] grid. The median value is reported across the dataset.  
(3) \emph{Angular (Conformal) Distortion}: We compute the cosine between the tangent and bitangent vectors of each face. The distortion for a shape is defined as one minus the average cosine across all faces~\cite{srinivasan2024nuvo}. We report the average distortion across the dataset. 
(4) \emph{Area (Equiareal) Distortion}: We compute both \emph{area distortion} and \emph{overall area distortion}. \emph{Area distortion} is defined as in Equation~\ref{equ:distortion}, based on the chart with the highest distortion in each shape. \emph{Overall area distortion} is computed by aggregating all triangles across all charts, calculating individual distortions, and averaging them. For both metrics, we first compute per-shape values and then average them across the dataset. All triangle-level distortion values are clipped to a maximum of 10. Notably, \emph{overall area distortion} may be smoothed by the number of faces, while \emph{area distortion} more effectively highlights problematic regions in the UV maps. We did not report stretch $L_2$ and $L_\infty$~\cite{sander2001texture} since they are per-triangle metrics that can diverge to infinity when triangles flip or have near-zero area---issues that are common in baselines such as xatlas, making direct comparison less meaningful. 

\begin{figure}[t]
    \centering
    \includegraphics[width=0.94\linewidth]{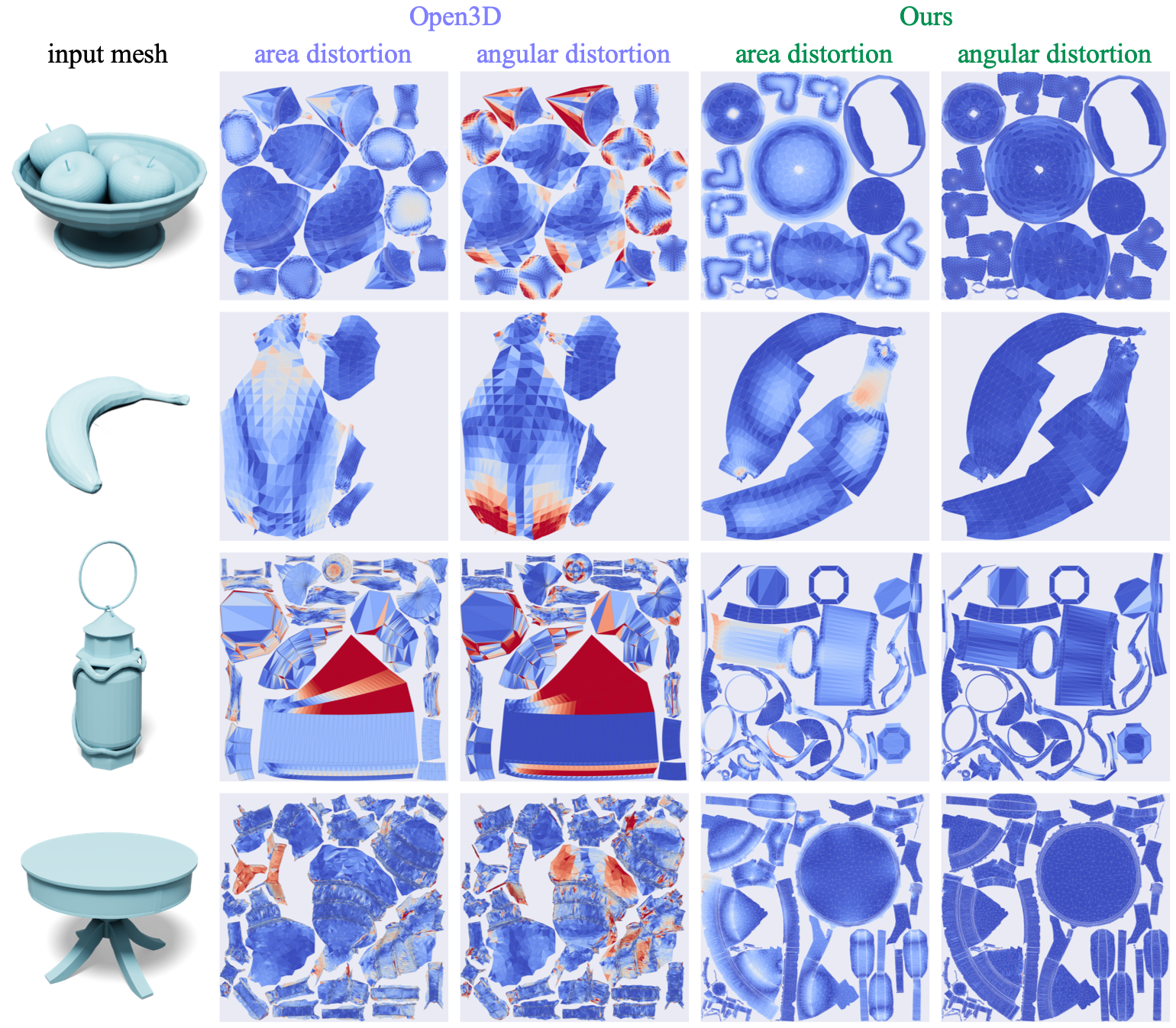}
       \vspace{-3ex}
    \captionof{figure}{Open3D not only suffers from a limited success rate but also produces results with large distortion, which may reduce their practical utility.}
    \label{fig:open3d}
    \vspace{-1em}
\end{figure}

\begin{figure}[t]
    \centering
    \includegraphics[width=0.94\linewidth]{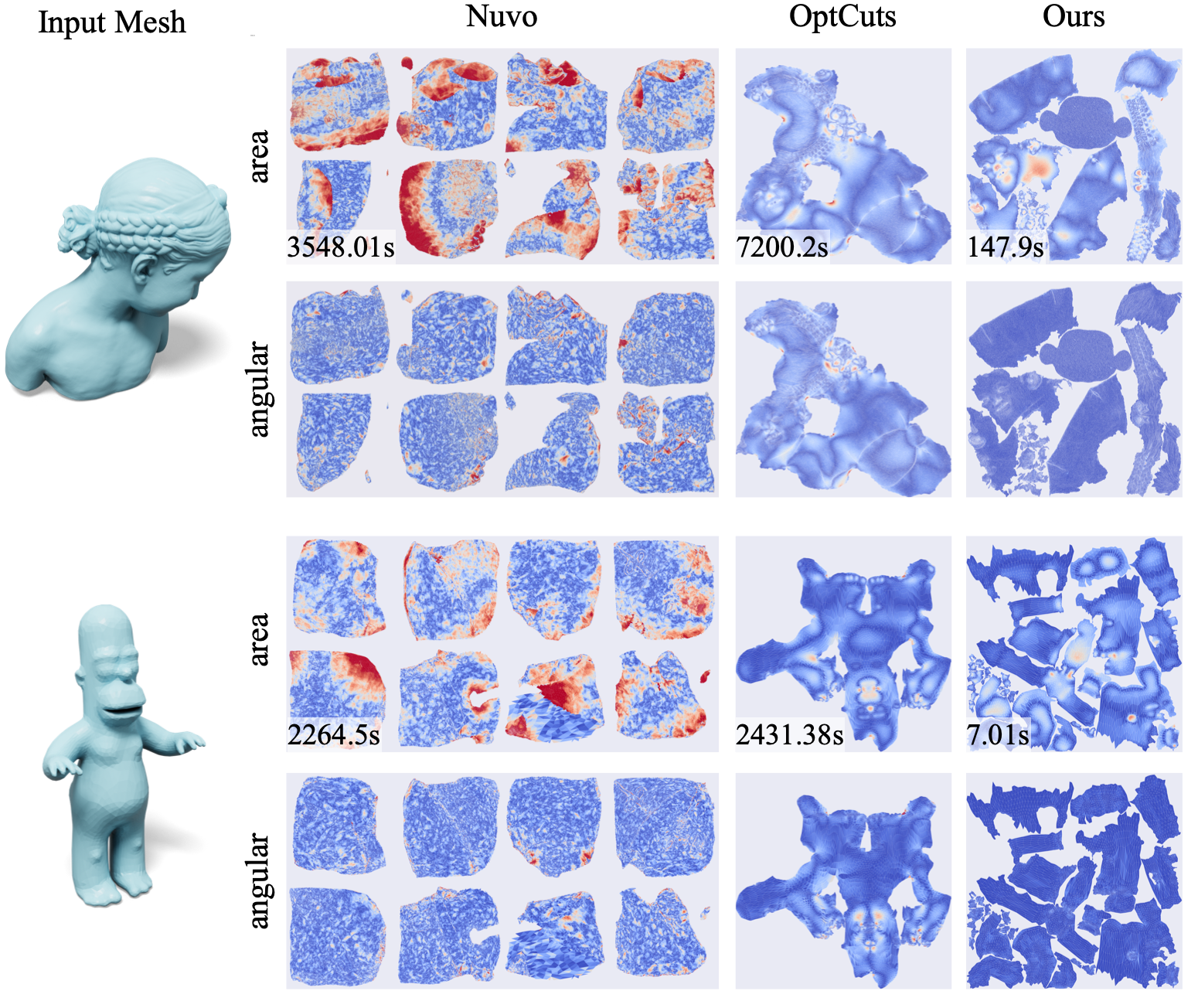}
   \vspace{-3ex}
    \captionof{figure}{Nuvo and OptCuts incur significantly longer optimisation times for each shape. Nuvo’s results also exhibit large distortion.}
    \label{fig:nuvo_optcut}
\end{figure}

\subsection{Comparison with Baselines}

\noindent \textbf{\textbf{Baselines.}}
We compare PartUV with commonly used tools—Blender’s Smart UV Project~\cite{blender}, xatlas~\cite{xatlas}, and Open3D—as well as optimization-based methods, including Nuvo~\cite{srinivasan2024nuvo} and OptCuts~\cite{li2018optcuts}.

Blender, xatlas, and Open3D decompose meshes into charts using bottom-up strategies guided by local geometric priors. Blender clusters triangles based on mesh normals and flattens each chart using simple planar projection. xatlas employs a greedy algorithm that balances geometric deviation, UV distortion, and seam cost, followed by Least Squares Conformal Maps (LSCM)~\cite{lscm} for flattening. Open3D builds on Microsoft’s UVAtlas~\cite{UVAtlas,zhou2004iso,sander2002signal}, which uses region growing guided by the isometric stretch metric and flattens charts with LSCM. Among optimization-based methods, Nuvo~\cite{srinivasan2024nuvo} learns a continuous UV mapping via a neural field by minimizing distortion losses with various regularizations. OptCuts jointly optimizes surface cuts and distortion under a global bijectivity constraint. Both methods require significantly longer runtimes per mesh.

\begin{table}[t]
  \centering
  \scriptsize
  \setlength{\tabcolsep}{1.2pt}
  \caption{\textbf{Quantitative comparison between Blender~\cite{blender}, xatlas~\cite{xatlas}, and PartUV.} }
    \vspace{-1.5em}
    \begin{tabular}{c|c|cccccccc}
    
    \toprule
          & & \multicolumn{1}{c}{success} &    \multicolumn{1}{c}{average}   & \multicolumn{1}{c}{median}      & \multicolumn{1}{c}{median} &       & \multicolumn{1}{c}{ } & \multicolumn{1}{c}{overall  } & \\
          dataset &  method & \multicolumn{1}{c}{rate} & \multicolumn{1}{c}{\# of}  & \multicolumn{1}{c}{\# of} & \multicolumn{1}{c}{seam} &\multicolumn{1}{c}{angular}  & \multicolumn{1}{c}{area} & \multicolumn{1}{c}{area} &  \multicolumn{1}{c}{time} \\
          &       & \multicolumn{1}{c}{(\%)} & \multicolumn{1}{c}{ charts $\downarrow$} & \multicolumn{1}{c}{ charts $\downarrow$} & \multicolumn{1}{c}{ length $\downarrow$} & \multicolumn{1}{c}{ distort. $\uparrow$} & \multicolumn{1}{c}{distort. $\downarrow$} & \multicolumn{1}{c}{distort. $\downarrow$} & \multicolumn{1}{c}{ (s) } \\
    \midrule
    \multicolumn{1}{c|}{\multirow{3}[0]{*}{\parbox{1.5cm}{\centering Common Shapes\\(24 shapes)}}} 
          & Blender & 100.0 & 1360.3 & 332.5 & 44.7  & 0.906 & 1.172 & 1.102 & 0.3 \\
          & xatlas & 100.0 & 974.8 & 301.0 & 42.9  & 0.987 & 1.885 & 1.504 & 77.9 \\
          & ours  & 100.0 & \textbf{48.6} & \textbf{16.5} & \textbf{16.6} & 0.990 & 1.283 & 1.128 & 39.4 \\
    \midrule
    \multicolumn{1}{c|}{\multirow{3}[0]{*}{\parbox{1.5cm}{\centering Trellis \\ (114 shapes)}}} 
          & Blender & 100.0 & 3352.9 & 1957.0 & 94.5  & 0.921 & 1.252 & 1.107 & 1.1 \\
          & xatlas & 100.0 & 1541.6 & 895.0 & 91.2  & 0.984 & 2.357 & 1.093 & 13.1 \\
          & ours  & 100.0 & \textbf{568.7} & \textbf{233.5} & \textbf{58.8} & 0.981 & 1.373 & 1.145 & 26.0 \\
    \midrule
      \multicolumn{1}{c|}{\multirow{3}[0]{*}{\parbox{1.5cm}{\centering ABC \\ (100 shapes)}}} 
          & Blender & 100.0 & 305.3 & 78.0  & 25.0  & 0.992 & 1.122 & 1.067 & 0.7 \\
          & xatlas & 100.0 & 249.6 & 56.0  & 26.5  & 1.000 & 1.192 & 1.030 & 31.0 \\
          & ours  & 100.0 & \textbf{85.4} & \textbf{15.5} & \textbf{17.3} & 0.998 & 1.191 & 1.073 & 31.9 \\
    \midrule
     \multicolumn{1}{c|}{\multirow{3}[0]{*}{\parbox{1.5cm}{\centering Part\\Objaverse\\Tiny (200 shapes)}}} 
          & Blender & 100.0 & 1509.2 & 647.5 & 70.2  & 0.925 & 1.325 & 1.115 & 0.2 \\
          & xatlas & 100.0 & 1142.1 & 491.5 & 67.0  & 0.982 & 1.728 & 1.286 & 4.4 \\
          & ours  & 100.0 & \textbf{544.8} & \textbf{177.0} & \textbf{41.8} & 0.983 & 1.305 & 1.112 & 10.1 \\
    \bottomrule
    \end{tabular}%
  \label{tab:blender_and_xatlas}
\vspace{-1em}
\end{table}%

\begin{table}[t]
  \centering
  \scriptsize
  \setlength{\tabcolsep}{1.5pt}
  \caption{\textbf{Quantitative comparison between Open3d~\cite{Zhou2018} and PartUV.} Note that Open3D has a limited success rate, and the reported numbers are averaged over the easier cases it successfully completes. Despite this, Open3D still exhibits large distortion.}
  \vspace{-2em}
    \begin{tabular}{c|c|cccccccc}
    \toprule
          &       & \multicolumn{1}{c}{success}
                  & \multicolumn{1}{c}{average}
                  & \multicolumn{1}{c}{}
                  & \multicolumn{1}{c}{overall}
                  & \multicolumn{1}{c}{average}
                  & \multicolumn{1}{c}{median}
                  & \multicolumn{1}{c}{median}
                  &  \\
              \multicolumn{1}{c|}{dataset}     
              & \multicolumn{1}{c|}{method}
              & \multicolumn{1}{c}{rate}
                  & \multicolumn{1}{c}{angular}
                  & \multicolumn{1}{c}{area}
                  & \multicolumn{1}{c}{area}
                  & \multicolumn{1}{c}{\# of}
                  & \multicolumn{1}{c}{\# of}
                  & \multicolumn{1}{c}{seam}
                  & \multicolumn{1}{c}{time} \\
        &
        & \multicolumn{1}{c}{(\%)}
        & \multicolumn{1}{c}{ distort. $\uparrow$}
        & \multicolumn{1}{c}{distort. $\downarrow$}
        & \multicolumn{1}{c}{distort. $\downarrow$}
        & \multicolumn{1}{c}{charts $\downarrow$}
        & \multicolumn{1}{c}{charts $\downarrow$}
        & \multicolumn{1}{c}{ length $\downarrow$}
        & \multicolumn{1}{c}{ (s) } \\
    \midrule
 \multicolumn{1}{c|}{\multirow{2}[0]{*}{\parbox{1.5cm}{\centering Common Shapes}}} & Open3d & 79.2  & 0.852 & 1.509 & 1.191 & 19.8  & 12.0  & 11.8  & 19.8 \\
          & ours  & \textbf{100.0} & \textbf{0.987} & \textbf{1.281} & \textbf{1.128} & 24.4  & 12.0  & 15.1  & 52.3 \\
    \midrule
 \multicolumn{1}{c|}{\multirow{2}[0]{*}{\parbox{1.5cm}{\centering Trellis }}} & Open3d & 39.5  & 0.859 & 1.931 & 1.264 & 79.8  & 40.0  & 25.9  & 24.1 \\
          & ours  & \textbf{100.0} & \textbf{0.984} & \textbf{1.308} & \textbf{1.144} & 97.9  & 51.0  & 27.4  & 23.7 \\
    \midrule
 \multicolumn{1}{c|}{\multirow{2}[0]{*}{\parbox{1.5cm}{\centering ABC }}} & Open3d & 83.0  & 0.878 & 1.459 & 1.162 & 15.0  & 8.0   & 9.3   & 17.6 \\
          & ours  & \textbf{100.0} & \textbf{0.994} & \textbf{1.171} & \textbf{1.062} & 35.0  & 15.0  & 18.6  & 39.6 \\
    \midrule
 \multicolumn{1}{c|}{\multirow{2}[0]{*}{\parbox{1.5cm}{\centering PartObjaverse \\Tiny }}}  & Open3d & 52.5  & 0.799 & 2.772 & 1.295 & 161.5 & 80.0  & 25.6  & 10.0 \\
          & ours  & \textbf{100.0} & \textbf{0.957} & \textbf{1.254} & \textbf{1.117} & 227.1 & 91.0  & 30.6  & 14.0 \\
    \bottomrule
    \end{tabular}%
  \label{tab:Open3d}%
  \vspace{-1em}
\end{table}%

\begin{table}[t]
  \centering
  \scriptsize
  \setlength{\tabcolsep}{2.4pt}
  \caption{\textbf{Quantitative comparison between Nuvo~\cite{srinivasan2024nuvo} and PartUV.}}
  \vspace{-2em}
    \begin{tabular}{c|c|cccccc}
    \toprule
      &       & \multicolumn{1}{c}{success}
              & \multicolumn{1}{c}{average}
              & \multicolumn{1}{c}{}
              & \multicolumn{1}{c}{overall}
              & \multicolumn{1}{c}{average}
              &  \\
% — Second header row: sub‐labels / units —
              \multicolumn{1}{c|}{dataset}
              & \multicolumn{1}{c|}{method}
              & \multicolumn{1}{c}{rate}
                  & \multicolumn{1}{c}{angular}
                  & \multicolumn{1}{c}{area}
                  & \multicolumn{1}{c}{area}
                  & \multicolumn{1}{c}{\# of}
                  & \multicolumn{1}{c}{time} \\
        &
        & \multicolumn{1}{c}{(\%)}
        & \multicolumn{1}{c}{distortion $\uparrow$}
        & \multicolumn{1}{c}{distortion $\downarrow$}
        & \multicolumn{1}{c}{distortion $\downarrow$}
        & \multicolumn{1}{c}{charts $\downarrow$}
        & \multicolumn{1}{c}{ (s) } \\
    \midrule
    \multicolumn{1}{c|}{\multirow{2}[1]{*}{Common Shapes}} & nuvo  & 100.0 & 0.802 & 2.722 & \textcolor[rgb]{ .114,  .11,  .114}{1.940} & 17.0  & 2908.8 \\
          & ours  & \textbf{100.0} & \textbf{0.987} & \textbf{1.281} & \textbf{1.128} & 24.4  & 52.3 \\
    \bottomrule
    \end{tabular}%
  \label{tab:nuvo}%
\end{table}%

\noindent \textbf{Results.}
As shown in Table~\ref{tab:blender_and_xatlas}, Blender and xatlas often produce over-fragmented charts, whereas PartUV generates UV maps with significantly fewer charts. For instance, on the Common Shapes dataset, PartUV uses only 1/31 as many charts as Blender. Consequently, it also results in shorter seam lengths. PartUV maintains low levels of both angular and area distortion, while xatlas may exhibit large area distortion on certain challenging shapes. Despite utilizing a more exhaustive decomposition search and a computationally expensive ABF flattening algorithm to achieve higher quality, PartUV maintains a runtime comparable to xatlas, typically completing in tens of seconds. See Figure~\ref{fig:partuv_comparison} for a qualitative comparison, where PartUV leverages semantic part information to produce chart boundaries that align more closely with object semantics.

\begin{figure}[t]
    \centering
    \includegraphics[width=0.98\linewidth]{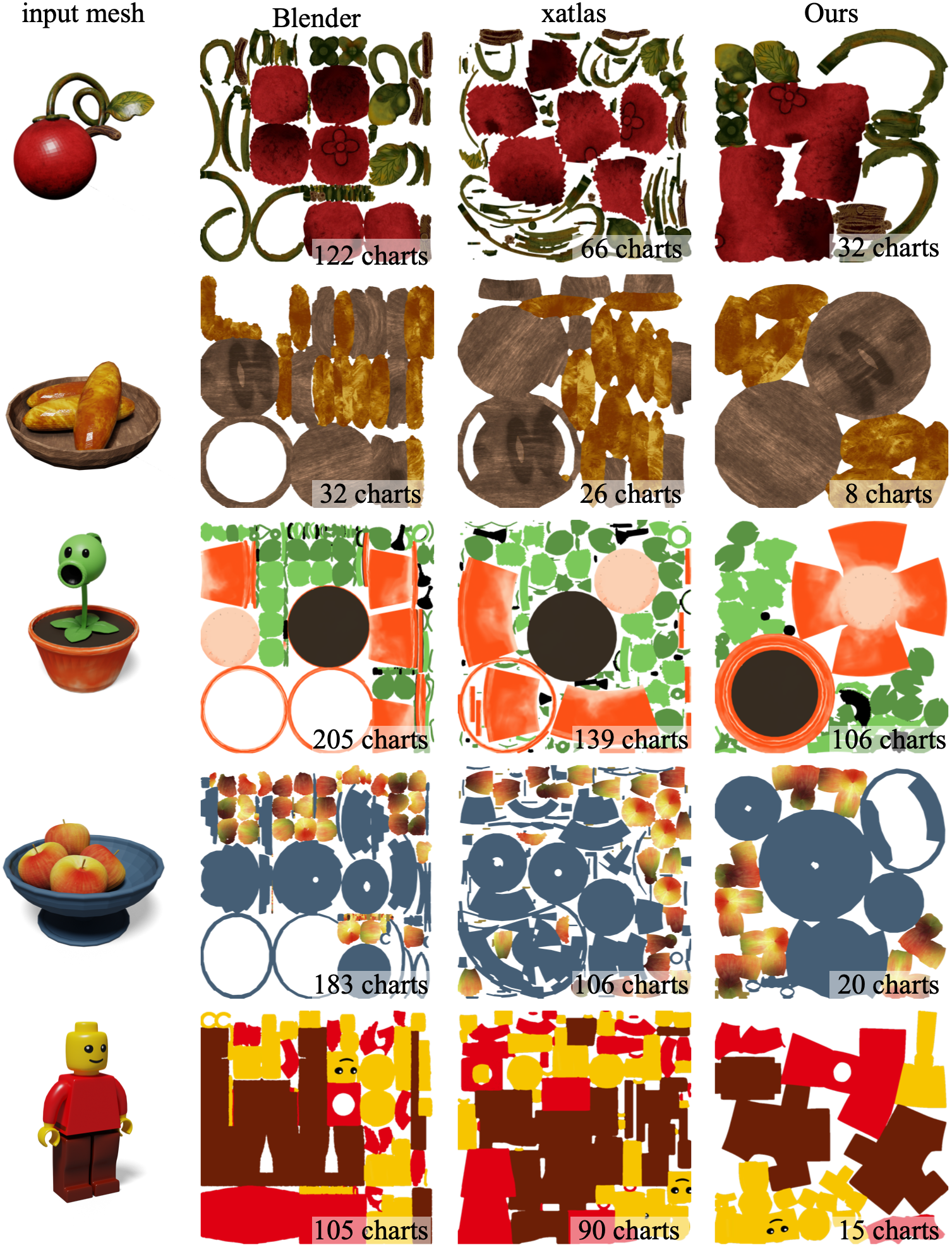}
    \vspace{-1em}
    \captionof{figure}{Texture-map comparison among Blender, xatlas, and our PartUV.}
    
    \label{fig:baking}
\end{figure}

We report the comparison results with Open3D separately in Table~\ref{tab:Open3d} due to its failure to complete some shapes within a reasonable time. 
For example, on the challenging Trellis dataset, it achieves a success rate of only 39.5\%. In Table~\ref{tab:Open3d}, we report the average performance only over the easy cases that Open3D successfully processes. For these cases, Open3D achieves a similar number of charts and seam lengths compared to our method. However, this comes at the cost of significant distortion. For instance, while Blender, xatlas, and PartUV all achieve angular distortion scores mostly above 0.95 across all datasets, Open3D consistently falls below 0.9 and even 0.8 in some cases. A similar phenomenon is observed for area distortion. Please refer to Figure~\ref{fig:open3d} and Figure~\ref{fig:baking} for qualitative examples, where Open3D produces large distortions, rendering the UV mappings less suitable for practical applications.

Compared to Nuvo and OptCuts, we observe that although both methods effectively reduce the number of charts, they typically require significantly longer optimization times—often exceeding 30 minutes or even several hours. Moreover, Nuvo leads to substantially higher distortion, as shown in Table~\ref{tab:nuvo} and Figure~\ref{fig:nuvo_optcut}. While OptCuts achieves low distortion, its success rate is limited: it produces outputs for only 9 out of 24 shapes in the Common Shapes dataset. Additionally, neither method incorporates the concept of semantic parts during optimization.

See supplementary material for additional qualitative examples UV efficiency comparisons, and analysis of Open3D results.

\subsection{Applications and Analysis}

\begin{figure}[t]
    \centering
    \includegraphics[width=\linewidth]{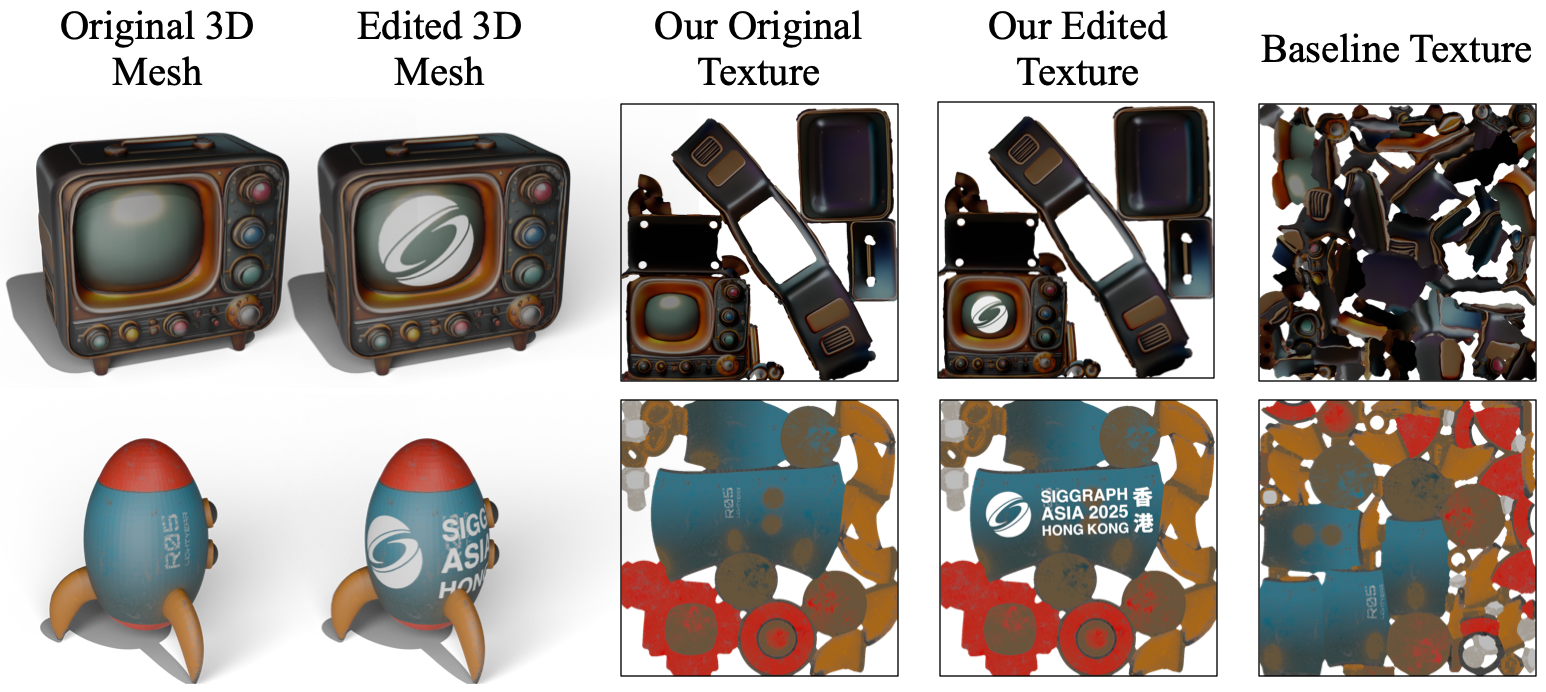} 
    \vspace{-2.5em}
    \caption{Unlike our baselines, which generate over-fragmented UV maps that hinder 2D texture editing, PartUV produces significantly fewer charts with part-aligned boundaries, enabling more effective 2D operations.} 
    \vspace{-1em}
    \label{fig:partuv_logo} 
\end{figure}

\begin{figure}[t]
  \centering
  
  \includegraphics[width=0.94\linewidth]{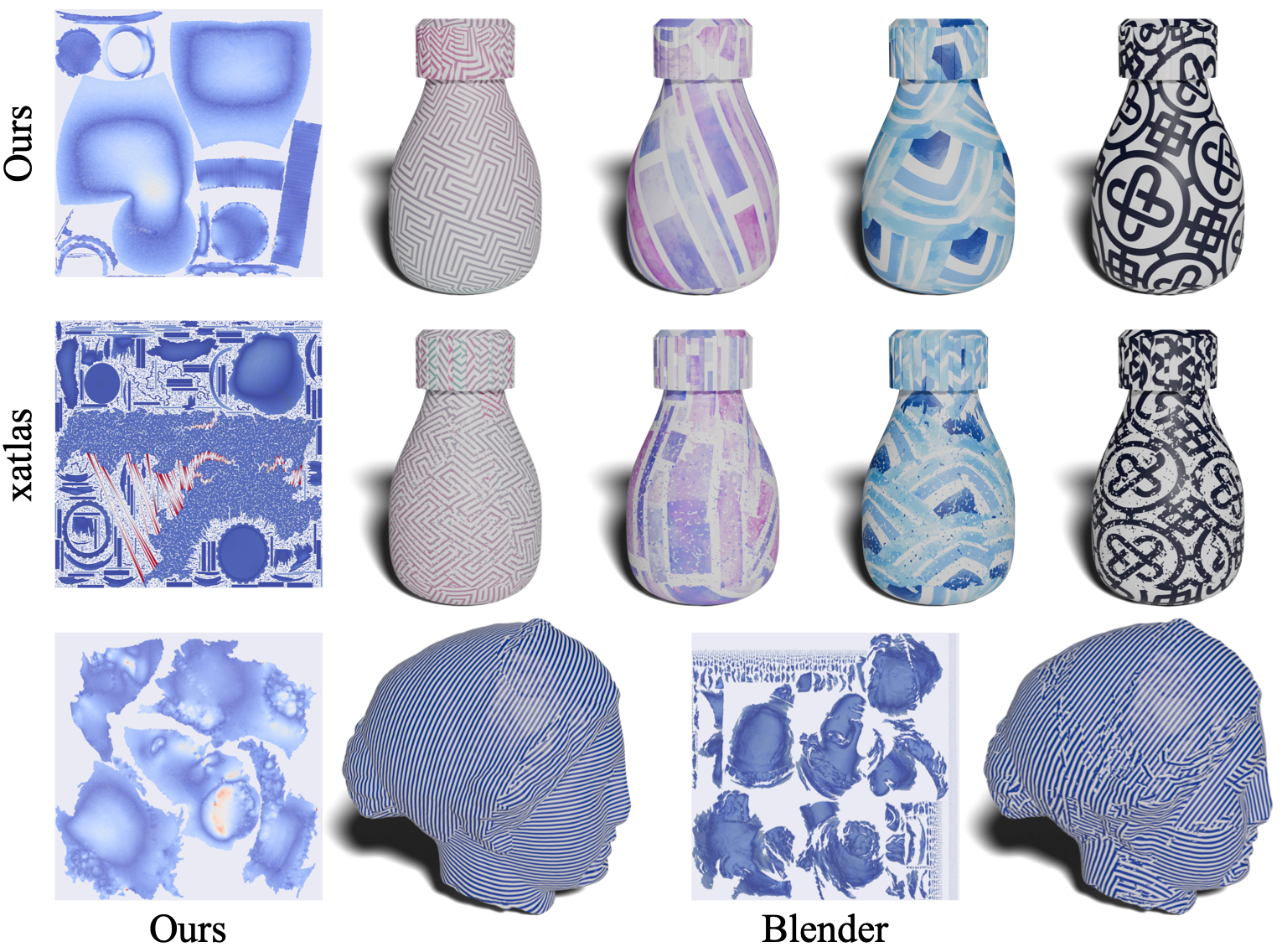}
  \vspace{-1em}
  \captionof{figure}{Our UV map enables easy texture replacement, whereas xatlas / Blender maps cause severe artefacts due to over-segmentation.}
  \label{fig:texture}
\end{figure}

In this section, we demonstrate the benefits of using our PartUV:

\noindent \textbf{Texture Editing and Replacement.}  
Since our UV mappings are less fragmented, texture maps can be more effectively edited or modified in 2D space. In Figure~\ref{fig:partuv_logo}, we show that our UV maps enable clean placement of conference logos, whereas the UV maps generated by Blender or xatlas fail to do so due to their overly fragmented layouts. Figure~\ref{fig:texture} further showcases an application where the texture is replaced with various tiling patterns. Because xatlas and Blender often produce numerous small charts containing only a few triangles, noticeable artifacts appear on the mesh surface. In contrast, our method preserves significantly better visual quality.

\begin{figure}[t]
    \centering
    \includegraphics[width=\linewidth]{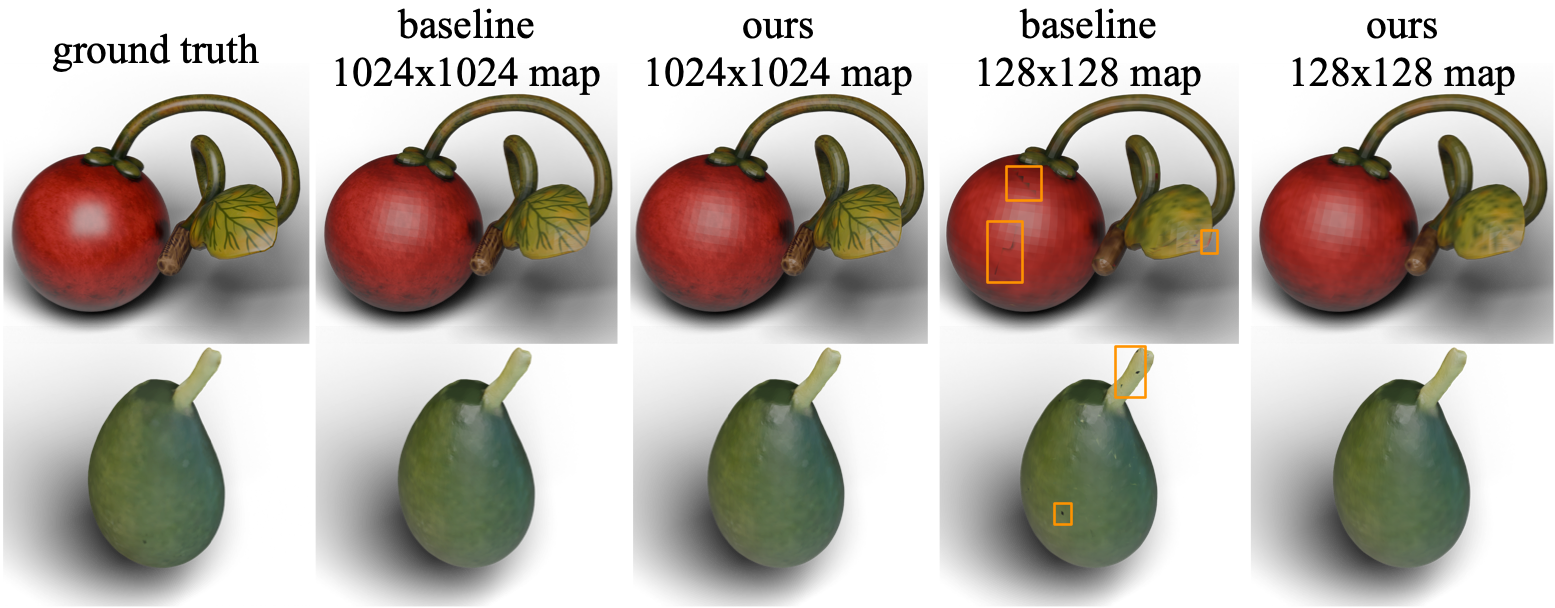} % Ensure this path is correct
    \vspace{-2em}
    \caption{xatlas and Blender generate over-fragmented UV maps that may introduce color bleeding, especially with low-resolution texture maps (e.g., in mobile games). In contrast, our results are free from such issues.} 
    \vspace{-0.5em}
    \label{fig:partuv_bleeding} 
\end{figure}

\noindent \textbf{Texture Compression.}  
UV maps always require padding between charts. When the UV layout is over-segmented, more padding is needed, which increases the risk of color bleeding. In Figure~\ref{fig:partuv_bleeding}, we demonstrate that reducing the UV map resolution from 1024 to 128—a common setting in mobile games—results in noticeable color bleeding for textures generated by xatlas and Blender. In contrast, PartUV is free from such issues.

\begin{figure*}[t]
    \centering
    \includegraphics[width=\linewidth]{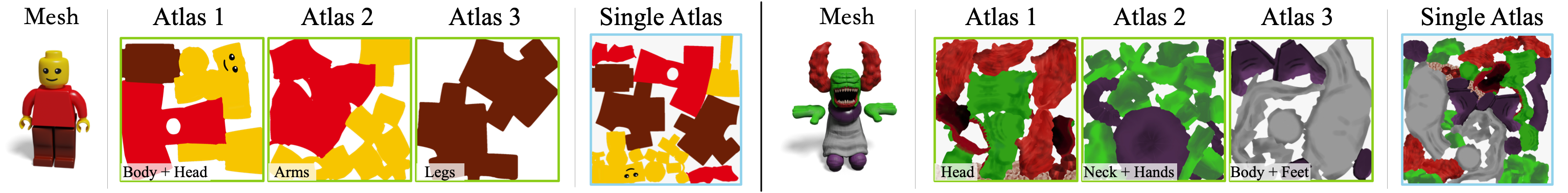}
    \vspace{-2.5em}
    \captionof{figure}{Our part-based UV unwrapping can pack all charts into a single atlas or multiple atlases by semantic part, aiding tasks such as 2-D editing.}
    \vspace{-1em}
    \label{fig:multiatlas}
\end{figure*}

\noindent \textbf{Multi-Atlas Wrapping.}  
As shown in Figure~\ref{fig:teaser} and Figure~\ref{fig:multiatlas}, PartUV supports part-based UV packing. Given the desired number of atlases, it can automatically extract semantic-meaningful parts and pack across separate atlases, facilitating downstream applications such as texture editing. 

\begin{figure}
      \centering
      \includegraphics[width=\linewidth]{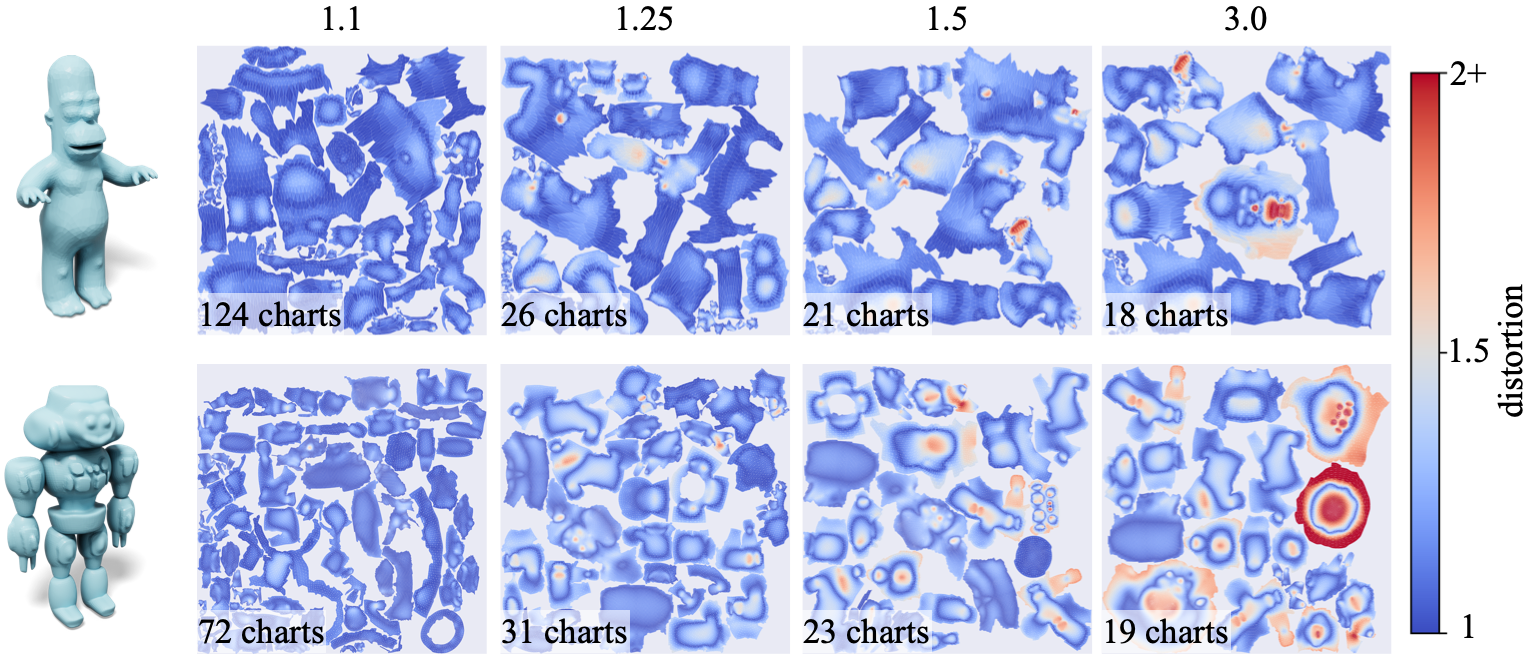}
    \vspace{-5ex}
      
      \captionof{figure}{PartUV lets users set a distortion threshold, flexibly controlling both distortion and chart count.}
      \vspace{-0.5em}
      \label{fig:threshold}
\end{figure}

\noindent \textbf{Adaptive Threshold Adjustment.}  
PartUV allows users to specify a distortion threshold $\tau$, enabling adaptive control over the trade-off between the number of charts and the distortion in the generated UV maps, as shown in Figure~\ref{fig:threshold}.

\subsection{Ablation Studies}

\noindent\textbf{PartField.} We integrate semantic part priors with geometric heuristics through an adaptive recursive tree search. A naive way to combine them is to first use PartField to decompose the shape into a fixed set of 20 parts, and then further decompose each part using the two heuristics. As shown in Table~\ref{tab:ablation}(a), this naive combination results in high distortion, as some parts may remain too complex to be flattened with low distortion. We also experimented with replacing the PartField features with face normals and then applying the original recursive tree search. As shown in Table~\ref{tab:ablation}(b), this variant doubles the runtime, increases the number of charts, and produces chart boundaries that no longer align with semantic parts.

\noindent\textbf{Merge Heuristic.} We propose an innovative geometry-based heuristic, \emph{Merge}, for chart decomposition. When this heuristic is removed and only the \emph{Normal} heuristic is used (Table~\ref{tab:ablation}(c)), we observe an increase in the number of charts.

\noindent\textbf{Recursion.} As shown in Alg.~\ref{alg:search}, Line~\ref{line:recursion}, we do not immediately return the first solution found but continue the search to find potentially better solutions. When this strategy is removed (Table~\ref{tab:ablation}(d)), we observe a significant increase in the number of charts.

\noindent\textbf{Distortion Surrogate.} To accelerate the search, we first simplify the mesh during intermediate iterations and then apply the ABF algorithm to the simplified mesh to compute an approximate distortion, which helps guide the search more efficiently. When this strategy is removed (Table~\ref{tab:ablation}(e)), we observe an increase in runtime.

We show more ablations and visualizations, including comparisons of different flattening algorithms, in the Appendix.
% \begin{table}[t]
%   \centering
%   \scriptsize
%   \setlength{\tabcolsep}{1pt}
%   \vspace{-1em}
%   \caption{\textbf{Ablation study conducted on the Trellis dataset.}}
%   \vspace{-2em}
%     \begin{tabular}{c|c|rrrrrr}
%     \toprule
%    &  \multirow{2}[2]{*}{version} & \multicolumn{1}{c}{average} & \multicolumn{1}{c}{median} & \multicolumn{1}{c}{area} & \multicolumn{1}{c}{angular} & \multicolumn{1}{c}{seam} & \multicolumn{1}{c}{\multirow{2}[2]{*}{time }} \\
%       id &    & \multicolumn{1}{c}{ \# charts $\downarrow$} & \multicolumn{1}{c}{ \# charts $\downarrow$} & \multicolumn{1}{c}{distort.$\downarrow$} & \multicolumn{1}{c}{distort.$\uparrow$} & \multicolumn{1}{c}{  length $\downarrow$} &  \\
%     \midrule
%     a & fixed 20 parts & 397.02 & 223.00 & 2.18  & 0.9687 & 66.42 & 207.75 \\
%     b & replace PF feat. with face normal & 574.43 & 259.50 & 1.28  & 0.9708 & 64.46 & 83.67 \\
%     c & no merge & 763.74 & 270.50 & 1.30  & 0.9607 & 62.75 & 38.00 \\
%     d & no recursion & 928.23 & 237.00 & 1.31  & 0.9630 & 59.17 & 40.66 \\
%     e & no distortion surrogate & 575.31 & 216.50 & 1.24  & 0.9632 &  57.88 & 61.48 \\
%     f & full  & 538.81 & 221.50 & 1.30  & 0.9609 & 57.92 & 41.88 \\
%     \bottomrule
%     \end{tabular}%
%     \vspace{-0.5em}
%   \label{tab:ablation}%
% \end{table}%

\begin{table}[t]
  \centering
  \scriptsize
  \setlength{\tabcolsep}{1pt}
  \vspace{-1em}
  
  \caption{\textbf{Ablation study conducted on the Trellis dataset.}}
  \vspace{-2em}
    \begin{tabular}{c|c|rrrrrr}
    \toprule
   &  \multirow{2}[2]{*}{version} & \multicolumn{1}{c}{average} & \multicolumn{1}{c}{median} & \multicolumn{1}{c}{area} & \multicolumn{1}{c}{angular} & \multicolumn{1}{c}{seam} & \multicolumn{1}{c}{\multirow{2}[2]{*}{time }} \\
      id &    & \multicolumn{1}{c}{ \# charts $\downarrow$} & \multicolumn{1}{c}{ \# charts $\downarrow$} & \multicolumn{1}{c}{distort.$\downarrow$} & \multicolumn{1}{c}{distort.$\uparrow$} & \multicolumn{1}{c}{  length $\downarrow$} &  \\
    \midrule
    a & fixed 20 parts & 397.02 & 223.00 & 2.18  & 0.9687 & 66.42 & 207.75 \\
    b & replace PF feat. with face normal & 574.43 & 259.50 & 1.28  & 0.9708 & 64.46 & 83.67 \\
    c & no merge & 778.64 & 299.50 & 1.42  & 0.9815 & 63.25 & 24.59 \\
    d & no recursion & 633.29 & 255.50 & 1.35  & 0.9822 & 61.02 & 33.82 \\
    e & no distortion surrogate & 580.52 & 248.00 & 1.25  & 0.9822 & 60.98 & 32.88 \\
    f & full  & 568.75 & 233.50 & 1.37  & 0.9811 & 59.85 & 26.03 \\
    \bottomrule
    
    \end{tabular}%
    \vspace{-0.5em}
  \label{tab:ablation}%
\end{table}%

\subsection{95th Percentile Distortion Metrics}
To evaluate worst-case behavior, we additionally report 95th-percentile distortion metrics and compare them with baseline methods on four datasets. As shown in Table~\ref{tab:95percentile}, we consider two metrics: \emph{distortion 95th-shape}, defined as the 95th-percentile area distortion across all shapes in a dataset, and \emph{distortion 95th-chart}, which first computes the 95th-percentile area distortion across all charts within a shape and then averages the results over all shapes in the dataset. As the table shows, our method consistently yields low 95th-percentile distortion values, with a maximum of 1.442, whereas baseline methods often produce much higher distortions (e.g., 4.701, 1.885, 1.728). These results demonstrate that our approach achieves more robust and stable performance under challenging cases.

\section{Discussion on Failure Cases}

Our method struggles with poor input mesh topology. For example, it cannot handle 3D meshes containing self-intersections. In such cases, the algorithm may recurse deeply in an attempt to resolve the intersections, which can lead to fragmented charts. Moreover, when input meshes are extremely fragmented—for instance, those with over 1000 components—the unwrapping results also become highly fragmented. In these cases, remeshing may be required before UV unwrapping. However, our method performs well on general meshes with few components and does not require meshes to be watertight or manifold.

\begin{table}[t]
\centering
\scriptsize
\setlength{\tabcolsep}{6pt}
\caption{\textbf{95th-percentile area distortion metrics across four datasets.}}
\vspace{-2em}
\begin{tabular}{c|c|ccc}
\toprule
\textbf{Dataset} & \textbf{Metric} & \textbf{Blender} & \textbf{xatlas} & \textbf{ours} \\
\midrule
\multicolumn{1}{c|}{\multirow{2}{*}{\textbf{ABC}}}
& distortion 95th-shape $\downarrow$    &  1.175 & 1.726 & 1.273 \\
\multicolumn{1}{c|}{}
& distortion 95th-chart $\downarrow$           & 1.093 & 1.041 & 1.133 \\
\midrule
\multicolumn{1}{c|}{\multirow{2}{*}{\textbf{Common Shapes}}}
& distortion 95th-shape $\downarrow$      & 1.885 & 1.504 & 1.404 \\
\multicolumn{1}{c|}{}
& distortion 95th-chart $\downarrow$           & 1.139 & 1.131 & 1.169 \\
\midrule
\multicolumn{1}{c|}{\multirow{2}{*}{\textbf{PartObjaverseTiny}}}
& distortion 95th-shape $\downarrow$      & 1.728 & 1.286 & 1.271 \\
\multicolumn{1}{c|}{}
& distortion 95th-chart $\downarrow$           & 1.132 & 1.079 & 1.116 \\
\midrule
\multicolumn{1}{c|}{\multirow{2}{*}{\textbf{Trellis}}}
& distortion 95th-shape $\downarrow$      & 1.319 & 4.701 & 1.442 \\
\multicolumn{1}{c|}{}
& distortion 95th-chart $\downarrow$           & 1.120 & 1.099 & 1.220 \\
\bottomrule
\end{tabular}
\label{tab:95percentile}
\vspace{-0.2em}
\end{table}

\vspace{1.5ex}
\section{Conclusion}
\vspace{-0.3em}

In this paper, we propose PartUV, a novel framework for UV unwrapping that strategically integrates semantic part priors from learning-based methods with two novel geometric heuristics. PartUV outperforms existing approaches by generating significantly fewer charts, low distortion, and chart boundaries that align with semantic parts. We demonstrate the advantages of this pipeline through several applications.

\clearpage
\twocolumn[{%
  {\sffamily\bfseries\Huge Appendix}
  \par
  \vspace{2em}
}]
\setcounter{section}{0}
\renewcommand{\thesection}{A\arabic{section}} 
\section{Additional Results and Comparisons}

\subsection{More Qualitative Results}

In the supplementary material, we provide an HTML file containing more qualitative and quantitative comparisons between our method and our baselines. The examples are gathered across the four datasets we used in the main paper. The metrics are computed and reported for individual shapes.

\subsection{Efficiency}

As we create the UV map with regard to semantic features and groups, it may raise doubts regarding the final efficiency of the resulting UV maps. In Table~\ref{tab:efficiency}, we show the comparison of the average efficiency of valid results between our method and the baselines. In our experiments, we always group the charts from the same parts when packing, and we use UVPackMaster~\cite{uvpackmaster3} to get the final packed UV map. We set the "heuristic search" time to 3 seconds for UVPackMaster, while most meshes finish within microseconds. We define efficiency as the total area of valid 2D faces within the normalized $0\text{–}1$ UV space.
It can be seen that our method does not hurt the overall efficiency, and it remains competitive compared to our baseline methods.

\begin{figure}[t]
    \centering
    \includegraphics[width=\linewidth]{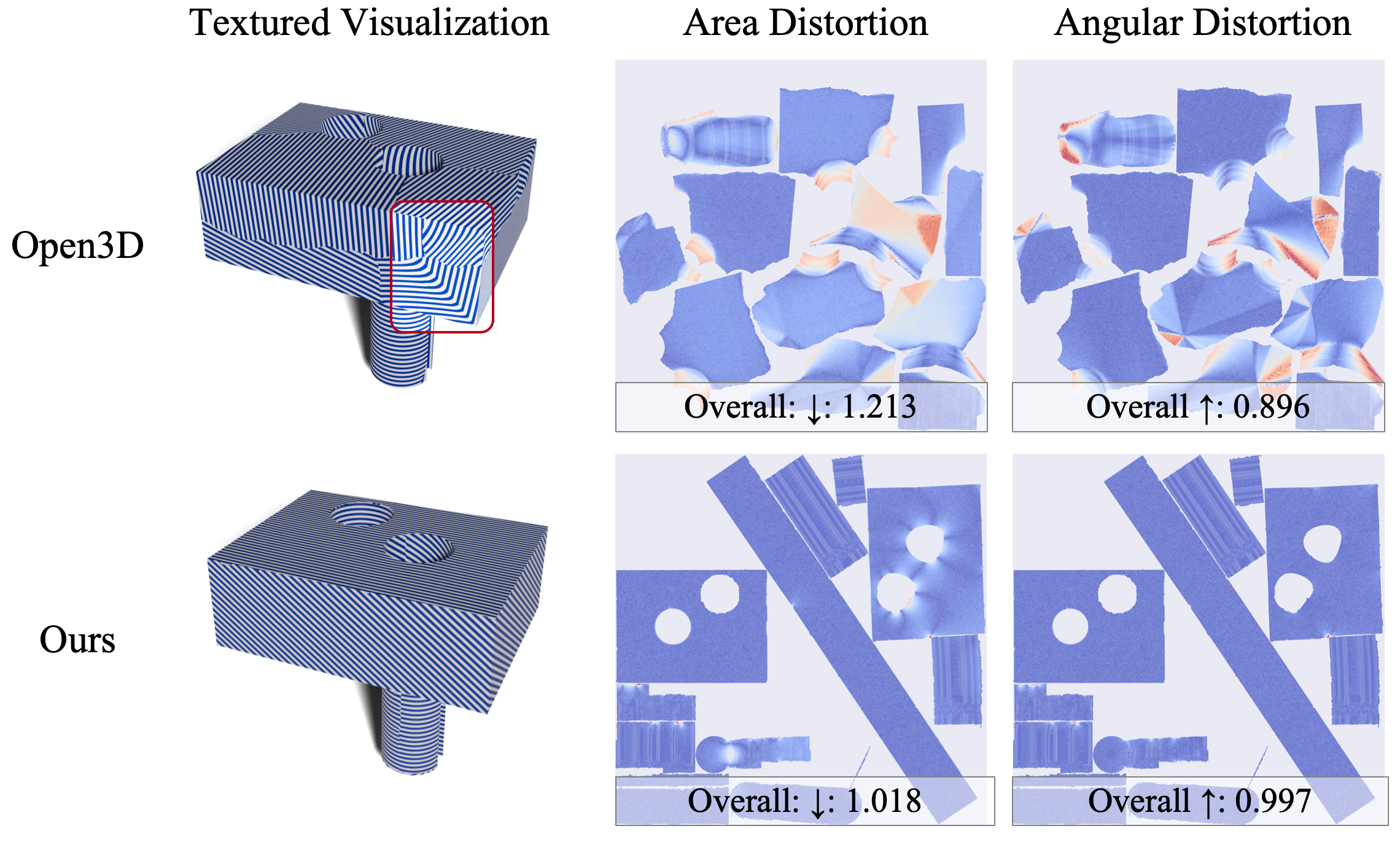}
    \caption{A demonstration of effects from Open3D's higher distortion results, creating curved (from angular distortion) and uneven (from area distortion) stripes. }
    \label{fig:o3d_stripe}
\end{figure}

\begin{table}
\centering
\caption{Efficiency ($\uparrow$) comparison across methods}
\label{tab:efficiency}
\begin{tabular}{lrrrr}
\toprule
 & ours & Blender & Open3D & xatlas \\
\midrule
Part-Objaverse-Tiny & 0.606 & 0.514 & 0.332 & 0.666 \\
Trellis             & 0.571 & 0.430 & 0.242 & 0.619 \\
ABC                 & 0.568 & 0.554 & 0.490 & 0.646 \\
Common Shapes       & 0.575 & 0.460 & 0.461 & 0.561 \\
\bottomrule
\end{tabular}
\end{table}

\subsection{Analysis and Visualization of Open3D Results}
In our experiments, particularly on the more challenging meshes, Open3D frequently crashed or exceeded the allotted time budget. We designate a shape as timed-out when its runtime surpasses 30 seconds plus three times the longest runtime among other methods on the current shape. 

Even for the shapes where Open3D successfully produces a UV map, the results still exhibit notable shortcomings.
At first glance, Open3D seemingly generates results with chart numbers similar to our method, albeit with slightly higher distortion. However, such differences in distortion can significantly affect downstream tasks.
In Fig.~\ref{fig:o3d_stripe}, we show one example from the ABC dataset, which Open3D creates a layout with a similar chart number with ours. However, with 0.2 difference in area distortion and 0.1 difference in angular distortion, the deformation of the texture is clearly noticeable. In particular, when using a striped texture, angular distortion leads to curved lines (bottom half of highlighted region), while area distortion causes uneven width (top half of highlighted region)—both prominently visible in the red-marked rectangle of the figure, not to mention the seams that disregard underlying geometric features. In contrast, our results exhibit minimal distortion, maintain the regularity of the shape, and feature seams that largely respect the underlying geometric structure.

\section{Additional Ablation Results}

\subsection{Replacing PartField feature with Face Normals}

\begin{figure}[t]
    \centering
    \includegraphics[width=\linewidth]{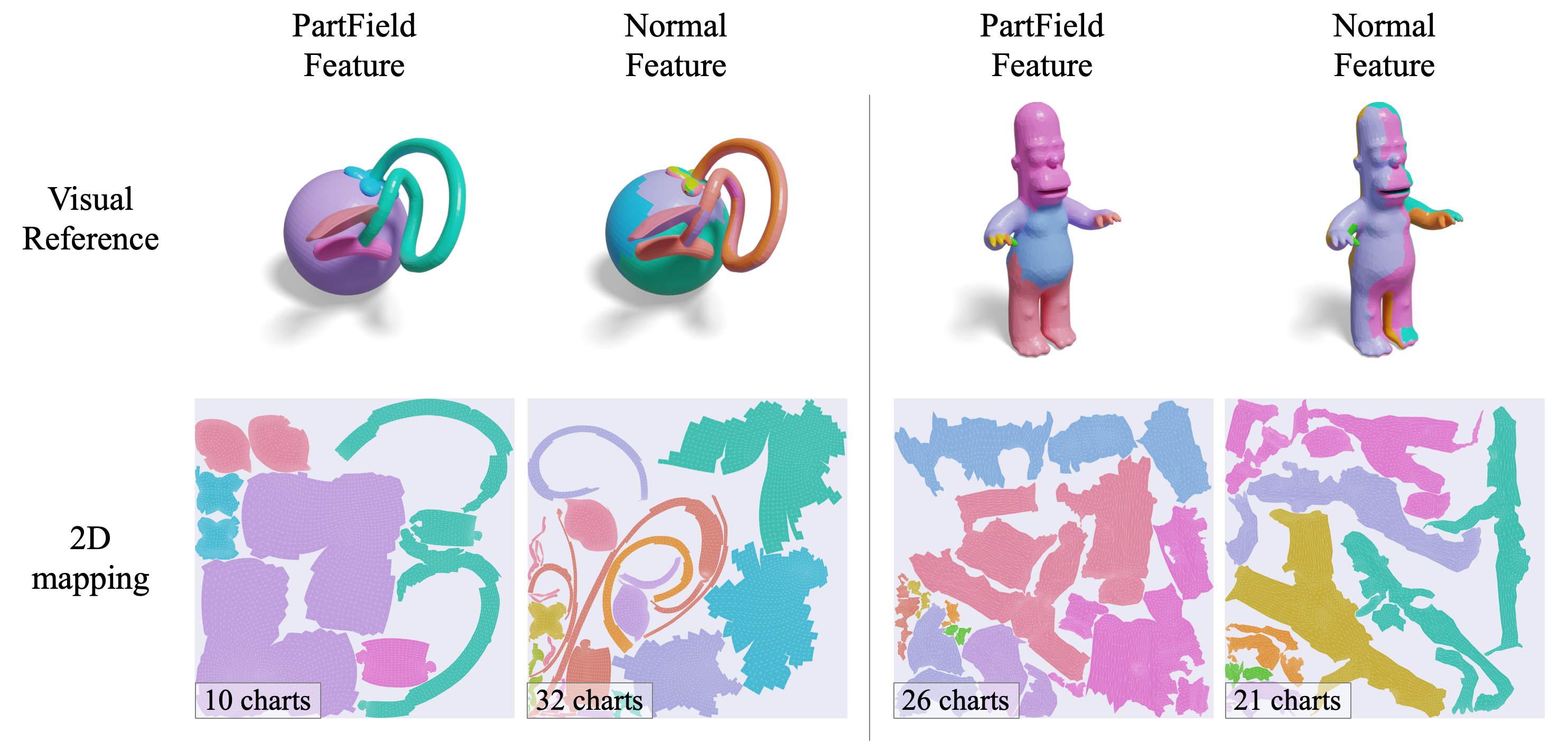}
    \caption{When using the face normals as agglomerative features for top-down tree construction, the pipeline generates messier results with no semantic alignment.}
    \label{fig:UNA}
\end{figure}

In our main pipeline, we create a top-down tree of faces with agglomerative clustering of PartField-predicted features. A simpler alternative would be replacing such features with face normals. In Fig.~\ref{fig:UNA}, we provide some visual examples of meshes using normal as Agglomerative features versus using the PartField-predicted ones. It can be seen that using normal is substantially prone to producing curly shapes, like the cherry on the left. Moreover, though being able to predict a similar number of charts under the same heuristic settings,  using normal as the agglomerative features leads to way less organized and neat UV layouts, as exemplified on the right of the figure.

\begin{figure}[t]
    \centering
    \includegraphics[width=\linewidth]{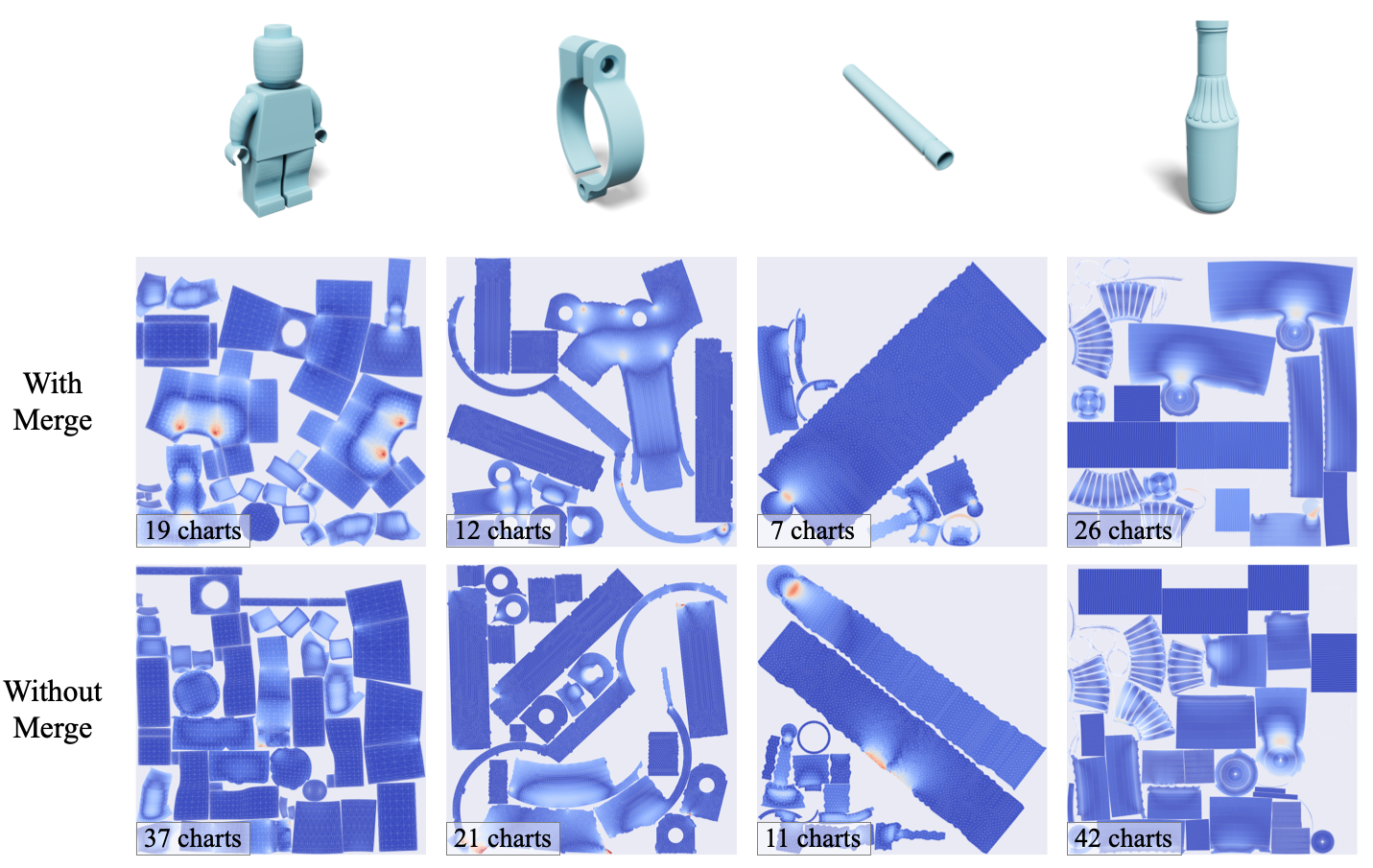}
    \caption{Visual comparison of with and without the merge heuristics.}
    \label{fig:nomerge}
\end{figure}

\begin{table}[t]
  \centering
  \small
  \setlength{\tabcolsep}{2pt}
  \caption{\textbf{Comparison between using ABF++ (Ours) and LSCM (Ours-LSCM) in our pipeline.}}

    \begin{tabular}{c|rrrrr}
    \toprule
        & \multicolumn{1}{c}{average} & \multicolumn{1}{c}{median} & \multicolumn{1}{c}{area} & \multicolumn{1}{c}{seam} & \multicolumn{1}{c}{time} \\
       & \multicolumn{1}{c}{\# charts $\downarrow$} & \multicolumn{1}{c}{\# charts $\downarrow$} & \multicolumn{1}{c}{distort.\,$\downarrow$} & \multicolumn{1}{c}{length $\downarrow$} & \multicolumn{1}{c}{(s)}   \\
    \midrule
     Ours-LSCM & 1154.12 & 363.00 & 1.27 & 65.48 & 58.61 \\
    Ours    &  538.81 & 221.50 & 1.30 & 57.92 & 41.88 \\
    \bottomrule
    \end{tabular}%
  \label{tab:lscm}%
\end{table}%

\subsection{Without Merge Heuristic}

In the main manuscript, we propose two heuristics used to unwrap a part into charts, \textit{Normal} and \textit{Merge}. Despite the higher cost of computation,  \textit{Merge} often produces more visually appealing UV maps with fewer charts. As illustrated in Fig.~\ref{fig:nomerge}, the \textit{Merge} heuristic can usually unwrap the part in an "unfolding" manner, creating less number of charts with a neater layout, comparing under the same distortion threshold requirements with using only the \textit{Normal} Heuristics. 

\subsection{Without Normal Heuristic }

\begin{figure}[t]
    \centering
    \includegraphics[width=\linewidth]{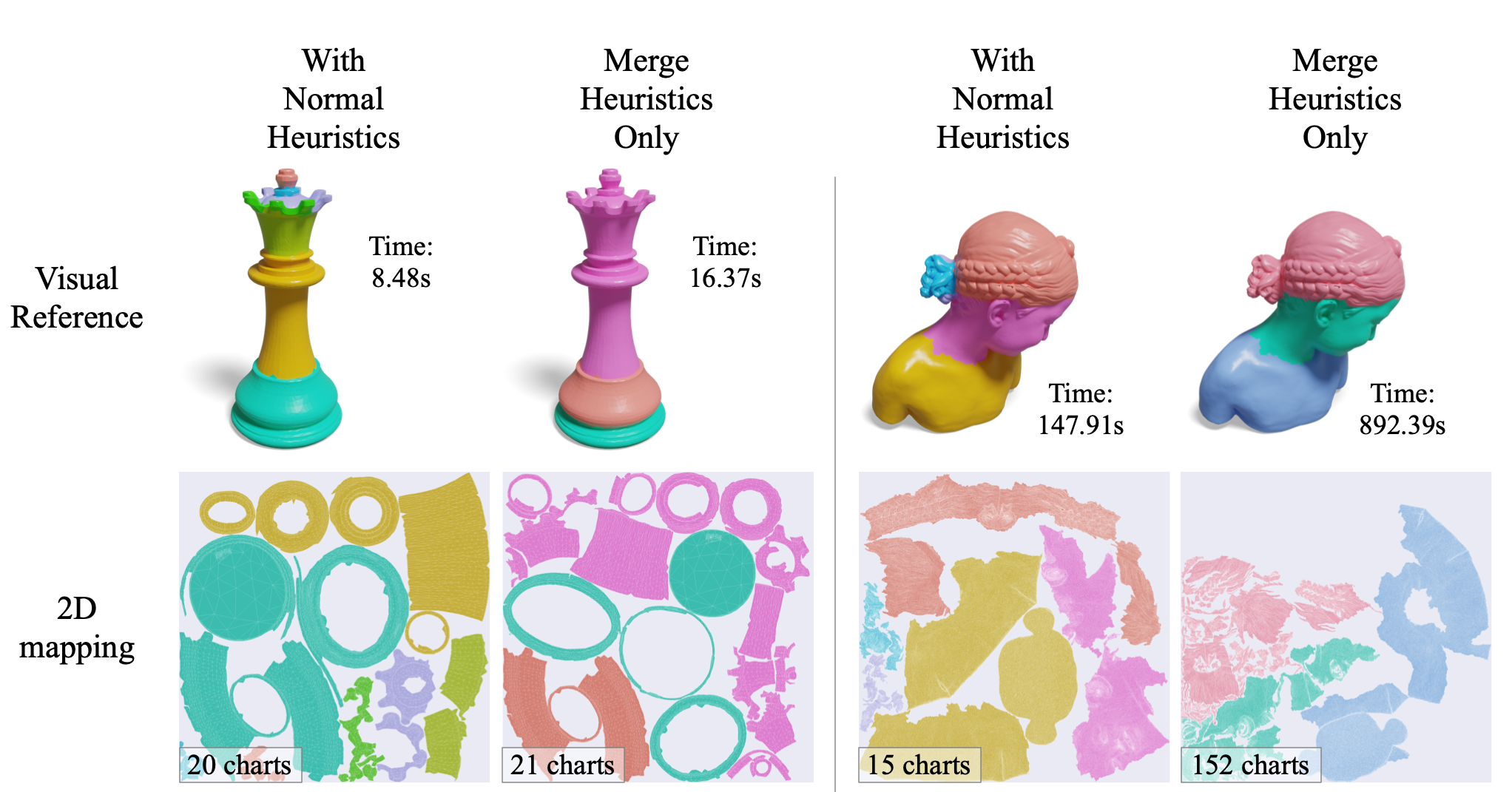}
    \caption{Using only \textit{Merge} heuristic from the beginning would incur more runtime, and more fragmented charts on bumpy meshes.}
    \label{fig:noAgg}
\end{figure}

With two heuristics included in the pipeline, one can also opt to use only the \textit{Merge} throughout the pipeline. We show several example results in Fig.~\ref{fig:noAgg}. Due to its high computational cost, using only \textit{Merge}, especially at the beginning, incurs substantial runtime overhead. To be precise, the \textit{Merge} heuristic would compute an oriented bounding box
(OBB) and use projection to get initial charts, and it could get hundreds of charts from a bumpy input. Trying to merge all of them would introduce significant overhead for such bumpy/uneven meshes. 
For example, on the bimba mesh on the right side of the figure, the runtime increases from 147 seconds to over 892 seconds. Moreover, due to its bumpy geometry, the merging process often results in overlapping, leading to an increase in the number of charts. In essence, \textit{Merge} could yield better results, but only on simpler shapes. Therefore, it is most effective as a complement to the \textit{Normal} heuristic and should be invoked only when \textit{Normal} produces a sufficiently good result.

\subsection{Using LSCM instead of ABF++ }

In our main pipeline, we adopt ABF++~\cite{sheffer2005abf++} as the primary flattening algorithm due to its fast performance and low-distortion mappings. Most of our baselines utilize LSCM~\cite{lscm}, which generally offers faster runtimes thanks to its linear energy formulation. However, despite its speed, LSCM generally produces inferior mapping results compared to ABF++. Such suboptimal outputs can trigger additional recursions in our pipeline, resulting in increased runtimes and more fragmented UV charts. As shown in Table~\ref{tab:lscm}, using LSCM ultimately leads to both higher runtime and a larger number of charts on the Trellis dataset.

\clearpage
\bibliographystyle{ACM-Reference-Format}
\bibliography{bib}


\begin{thebibliography}{82}

%%% the Web address.

\ifx \showCODEN    \undefined \def \showCODEN     #1{\unskip}     \fi
\ifx \showDOI      \undefined \def \showDOI       #1{#1}\fi
\ifx \showISBNx    \undefined \def \showISBNx     #1{\unskip}     \fi
\ifx \showISBNxiii \undefined \def \showISBNxiii  #1{\unskip}     \fi
\ifx \showISSN     \undefined \def \showISSN      #1{\unskip}     \fi
\ifx \showLCCN     \undefined \def \showLCCN      #1{\unskip}     \fi
\ifx \shownote     \undefined \def \shownote      #1{#1}          \fi
\ifx \showarticletitle \undefined \def \showarticletitle #1{#1}   \fi
\ifx \showURL      \undefined \def \showURL       {\relax}        \fi
% invisible to TeX
\providecommand\bibfield[2]{#2}
\providecommand\bibinfo[2]{#2}
\providecommand\natexlab[1]{#1}
\providecommand\showeprint[2][]{arXiv:#2}

\bibitem[\protect\citeauthoryear{Alexa, Cohen-Or, and Levin}{Alexa
  et~al\mbox{.}}{2023}]%
        {alexa2023rigid}
\bibfield{author}{\bibinfo{person}{Marc Alexa}, \bibinfo{person}{Daniel
  Cohen-Or}, {and} \bibinfo{person}{David Levin}.}
  \bibinfo{year}{2023}\natexlab{}.
\newblock \showarticletitle{As-rigid-as-possible shape interpolation}.
\newblock In \bibinfo{booktitle}{\emph{Seminal Graphics Papers: Pushing the
  Boundaries, Volume 2}}. \bibinfo{pages}{165--172}.
\newblock


\bibitem[\protect\citeauthoryear{Belkin and Niyogi}{Belkin and Niyogi}{2003}]%
        {belkin2003laplacian}
\bibfield{author}{\bibinfo{person}{Mikhail Belkin} {and}
  \bibinfo{person}{Partha Niyogi}.} \bibinfo{year}{2003}\natexlab{}.
\newblock \showarticletitle{Laplacian eigenmaps for dimensionality reduction
  and data representation}.
\newblock \bibinfo{journal}{\emph{Neural computation}} \bibinfo{volume}{15},
  \bibinfo{number}{6} (\bibinfo{year}{2003}), \bibinfo{pages}{1373--1396}.
\newblock


\bibitem[\protect\citeauthoryear{Ben-Chen, Gotsman, and Bunin}{Ben-Chen
  et~al\mbox{.}}{2008}]%
        {ben2008conformal}
\bibfield{author}{\bibinfo{person}{Mirela Ben-Chen}, \bibinfo{person}{Craig
  Gotsman}, {and} \bibinfo{person}{Guy Bunin}.}
  \bibinfo{year}{2008}\natexlab{}.
\newblock \showarticletitle{Conformal flattening by curvature prescription and
  metric scaling}. In \bibinfo{booktitle}{\emph{Computer Graphics Forum}},
  Vol.~\bibinfo{volume}{27}. Wiley Online Library, \bibinfo{pages}{449--458}.
\newblock


\bibitem[\protect\citeauthoryear{Bhargava, Schreck, Freire, Hugron, Lefebvre,
  Sell{\'a}n, and Bickel}{Bhargava et~al\mbox{.}}{2025}]%
        {bhargava2025mesh}
\bibfield{author}{\bibinfo{person}{Manas Bhargava}, \bibinfo{person}{Camille
  Schreck}, \bibinfo{person}{Marco Freire}, \bibinfo{person}{Pierre-Alexandre
  Hugron}, \bibinfo{person}{Sylvain Lefebvre}, \bibinfo{person}{Silvia
  Sell{\'a}n}, {and} \bibinfo{person}{Bernd Bickel}.}
  \bibinfo{year}{2025}\natexlab{}.
\newblock \showarticletitle{Mesh Simplification for Unfolding}. In
  \bibinfo{booktitle}{\emph{Computer Graphics Forum}},
  Vol.~\bibinfo{volume}{44}. Wiley Online Library, \bibinfo{pages}{e15269}.
\newblock


\bibitem[\protect\citeauthoryear{Carr, Hoberock, Crane, and Hart}{Carr
  et~al\mbox{.}}{2006}]%
        {carr2006rectangular}
\bibfield{author}{\bibinfo{person}{Nathan~A Carr}, \bibinfo{person}{Jared
  Hoberock}, \bibinfo{person}{Keenan Crane}, {and} \bibinfo{person}{John~C
  Hart}.} \bibinfo{year}{2006}\natexlab{}.
\newblock \showarticletitle{Rectangular multi-chart geometry images}. In
  \bibinfo{booktitle}{\emph{Symposium on geometry processing}}.
  \bibinfo{pages}{181--190}.
\newblock


\bibitem[\protect\citeauthoryear{Chen, Yin, and Fidler}{Chen
  et~al\mbox{.}}{2022}]%
        {chen2022auv}
\bibfield{author}{\bibinfo{person}{Zhiqin Chen}, \bibinfo{person}{Kangxue Yin},
  {and} \bibinfo{person}{Sanja Fidler}.} \bibinfo{year}{2022}\natexlab{}.
\newblock \showarticletitle{Auv-net: Learning aligned uv maps for texture
  transfer and synthesis}. In \bibinfo{booktitle}{\emph{Proceedings of the
  IEEE/CVF conference on computer vision and pattern recognition}}.
  \bibinfo{pages}{1465--1474}.
\newblock


\bibitem[\protect\citeauthoryear{Community}{Community}{[n.d.]}]%
        {blender}
\bibfield{author}{\bibinfo{person}{Blender~Online Community}.}
  \bibinfo{year}{[n.d.]}\natexlab{}.
\newblock \bibinfo{title}{Blender - a {3D} modelling and rendering package}.
\newblock
\newblock
\newblock
\shownote{\url{http://www.blender.org}.}


\bibitem[\protect\citeauthoryear{Corporation}{Corporation}{2023}]%
        {UVAtlas}
\bibfield{author}{\bibinfo{person}{Microsoft Corporation}.}
  \bibinfo{year}{2023}\natexlab{}.
\newblock \bibinfo{title}{{UVAtlas}: Chart-based UV Atlas Generation Library}.
\newblock \bibinfo{howpublished}{\url{https://github.com/microsoft/UVAtlas}}.
\newblock
\newblock
\shownote{Accessed: 2025-05-21.}


\bibitem[\protect\citeauthoryear{Das, Ma, Shu, and Samaras}{Das
  et~al\mbox{.}}{2022}]%
        {LearninganIsometric}
\bibfield{author}{\bibinfo{person}{Sagnik Das}, \bibinfo{person}{Ke Ma},
  \bibinfo{person}{Zhixin Shu}, {and} \bibinfo{person}{Dimitris Samaras}.}
  \bibinfo{year}{2022}\natexlab{}.
\newblock \showarticletitle{Learning an Isometric Surface Parameterization
  for Texture Unwrapping}.
\newblock \bibinfo{journal}{\emph{European Conference on Computer Vision}}
  (\bibinfo{year}{2022}).
\newblock


\bibitem[\protect\citeauthoryear{Deitke, Schwenk, Salvador, Weihs, Michel,
  VanderBilt, Schmidt, Ehsani, Kembhavi, and Farhadi}{Deitke
  et~al\mbox{.}}{2022}]%
        {objaverse}
\bibfield{author}{\bibinfo{person}{Matt Deitke}, \bibinfo{person}{Dustin
  Schwenk}, \bibinfo{person}{Jordi Salvador}, \bibinfo{person}{Luca Weihs},
  \bibinfo{person}{Oscar Michel}, \bibinfo{person}{Eli VanderBilt},
  \bibinfo{person}{Ludwig Schmidt}, \bibinfo{person}{Kiana Ehsani},
  \bibinfo{person}{Aniruddha Kembhavi}, {and} \bibinfo{person}{Ali Farhadi}.}
  \bibinfo{year}{2022}\natexlab{}.
\newblock \showarticletitle{Objaverse: A Universe of Annotated 3D Objects}.
\newblock \bibinfo{journal}{\emph{arXiv preprint arXiv:2212.08051}}
  (\bibinfo{year}{2022}).
\newblock


\bibitem[\protect\citeauthoryear{Desbrun, Meyer, and Alliez}{Desbrun
  et~al\mbox{.}}{2002}]%
        {desbrun2002intrinsic}
\bibfield{author}{\bibinfo{person}{Mathieu Desbrun}, \bibinfo{person}{Mark
  Meyer}, {and} \bibinfo{person}{Pierre Alliez}.}
  \bibinfo{year}{2002}\natexlab{}.
\newblock \showarticletitle{Intrinsic parameterizations of surface meshes}. In
  \bibinfo{booktitle}{\emph{Computer graphics forum}},
  Vol.~\bibinfo{volume}{21}. Wiley Online Library, \bibinfo{pages}{209--218}.
\newblock


\bibitem[\protect\citeauthoryear{Gu, Gortler, and Hoppe}{Gu
  et~al\mbox{.}}{2002}]%
        {gu2002geometry}
\bibfield{author}{\bibinfo{person}{Xianfeng Gu}, \bibinfo{person}{Steven~J
  Gortler}, {and} \bibinfo{person}{Hugues Hoppe}.}
  \bibinfo{year}{2002}\natexlab{}.
\newblock \showarticletitle{Geometry images}. In
  \bibinfo{booktitle}{\emph{Proceedings of the 29th annual conference on
  Computer graphics and interactive techniques}}. \bibinfo{pages}{355--361}.
\newblock


\bibitem[\protect\citeauthoryear{Guo, Zhu, Peng, Wang, Shen, Hu, and Zhou}{Guo
  et~al\mbox{.}}{2024}]%
        {guo2024sam}
\bibfield{author}{\bibinfo{person}{Haoyu Guo}, \bibinfo{person}{He Zhu},
  \bibinfo{person}{Sida Peng}, \bibinfo{person}{Yuang Wang},
  \bibinfo{person}{Yujun Shen}, \bibinfo{person}{Ruizhen Hu}, {and}
  \bibinfo{person}{Xiaowei Zhou}.} \bibinfo{year}{2024}\natexlab{}.
\newblock \showarticletitle{Sam-guided graph cut for 3d instance segmentation}.
  In \bibinfo{booktitle}{\emph{European Conference on Computer Vision}}.
  Springer, \bibinfo{pages}{234--251}.
\newblock


\bibitem[\protect\citeauthoryear{He, Peng, Jiang, Hu, Zhang, Nie, Wang, and
  Wang}{He et~al\mbox{.}}{2024}]%
        {he2024pointseg}
\bibfield{author}{\bibinfo{person}{Qingdong He}, \bibinfo{person}{Jinlong
  Peng}, \bibinfo{person}{Zhengkai Jiang}, \bibinfo{person}{Xiaobin Hu},
  \bibinfo{person}{Jiangning Zhang}, \bibinfo{person}{Qiang Nie},
  \bibinfo{person}{Yabiao Wang}, {and} \bibinfo{person}{Chengjie Wang}.}
  \bibinfo{year}{2024}\natexlab{}.
\newblock \showarticletitle{PointSeg: A Training-Free Paradigm for 3D Scene
  Segmentation via Foundation Models}.
\newblock \bibinfo{journal}{\emph{arXiv preprint arXiv:2403.06403}}
  (\bibinfo{year}{2024}).
\newblock


\bibitem[\protect\citeauthoryear{Hormann and Greiner}{Hormann and
  Greiner}{2000}]%
        {hormann2000mips}
\bibfield{author}{\bibinfo{person}{Kai Hormann} {and}
  \bibinfo{person}{G{\"u}nther Greiner}.} \bibinfo{year}{2000}\natexlab{}.
\newblock \showarticletitle{MIPS: An efficient global parametrization method}.
\newblock \bibinfo{journal}{\emph{Curve and Surface Design: Saint-Malo 1999}}
  (\bibinfo{year}{2000}), \bibinfo{pages}{153--162}.
\newblock


\bibitem[\protect\citeauthoryear{Jacobson}{Jacobson}{2013}]%
        {jacobsonCommon3DTestModels}
\bibfield{author}{\bibinfo{person}{Alec Jacobson}.}
  \bibinfo{year}{2013}\natexlab{}.
\newblock \bibinfo{title}{common-3d-test-models}.
\newblock
  \bibinfo{howpublished}{\url{https://github.com/alecjacobson/common-3d-test-models}}.
\newblock
\newblock
\shownote{Accessed: 2025-05-21.}


\bibitem[\protect\citeauthoryear{Jacobson and contributors}{Jacobson and
  contributors}{2023}]%
        {jacobson_common3d_2023}
\bibfield{author}{\bibinfo{person}{Alec Jacobson} {and}
  \bibinfo{person}{contributors}.} \bibinfo{year}{2023}\natexlab{}.
\newblock \bibinfo{title}{{common-3d-test-models: Repository of common 3D test
  meshes}}.
\newblock
  \bibinfo{howpublished}{\url{https://github.com/alecjacobson/common-3d-test-models}}.
\newblock
\newblock
\shownote{Git commit 8a4f864, accessed 2025-05-15.}


\bibitem[\protect\citeauthoryear{Jiang, Zhao, Shi, Liu, Fu, and Jia}{Jiang
  et~al\mbox{.}}{2020}]%
        {jiang2020pointgroup}
\bibfield{author}{\bibinfo{person}{Li Jiang}, \bibinfo{person}{Hengshuang
  Zhao}, \bibinfo{person}{Shaoshuai Shi}, \bibinfo{person}{Shu Liu},
  \bibinfo{person}{Chi-Wing Fu}, {and} \bibinfo{person}{Jiaya Jia}.}
  \bibinfo{year}{2020}\natexlab{}.
\newblock \showarticletitle{Pointgroup: Dual-set point grouping for 3d instance
  segmentation}. In \bibinfo{booktitle}{\emph{Proceedings of the IEEE/CVF
  conference on computer vision and Pattern recognition}}.
  \bibinfo{pages}{4867--4876}.
\newblock


\bibitem[\protect\citeauthoryear{Johnson}{Johnson}{1967}]%
        {johnson1967hierarchical}
\bibfield{author}{\bibinfo{person}{Stephen~C Johnson}.}
  \bibinfo{year}{1967}\natexlab{}.
\newblock \showarticletitle{Hierarchical clustering schemes}.
\newblock \bibinfo{journal}{\emph{Psychometrika}} \bibinfo{volume}{32},
  \bibinfo{number}{3} (\bibinfo{year}{1967}), \bibinfo{pages}{241--254}.
\newblock


\bibitem[\protect\citeauthoryear{Julius, Kraevoy, and Sheffer}{Julius
  et~al\mbox{.}}{2005}]%
        {julius2005d}
\bibfield{author}{\bibinfo{person}{Dan Julius}, \bibinfo{person}{Vladislav
  Kraevoy}, {and} \bibinfo{person}{Alla Sheffer}.}
  \bibinfo{year}{2005}\natexlab{}.
\newblock \showarticletitle{D-charts: Quasi-developable mesh segmentation}. In
  \bibinfo{booktitle}{\emph{Computer Graphics Forum}},
  Vol.~\bibinfo{volume}{24}. Citeseer, \bibinfo{pages}{581--590}.
\newblock


\bibitem[\protect\citeauthoryear{Kalvin and Taylor}{Kalvin and Taylor}{1996}]%
        {kalvin1996superfaces}
\bibfield{author}{\bibinfo{person}{Alan~D Kalvin} {and}
  \bibinfo{person}{Russell~H Taylor}.} \bibinfo{year}{1996}\natexlab{}.
\newblock \showarticletitle{Superfaces: Polygonal mesh simplification with
  bounded error}.
\newblock \bibinfo{journal}{\emph{IEEE Computer Graphics and Applications}}
  \bibinfo{volume}{16}, \bibinfo{number}{3} (\bibinfo{year}{1996}),
  \bibinfo{pages}{64--77}.
\newblock


\bibitem[\protect\citeauthoryear{Katz and Tal}{Katz and Tal}{2003}]%
        {katz2003hierarchical}
\bibfield{author}{\bibinfo{person}{Sagi Katz} {and} \bibinfo{person}{Ayellet
  Tal}.} \bibinfo{year}{2003}\natexlab{}.
\newblock \showarticletitle{Hierarchical mesh decomposition using fuzzy
  clustering and cuts}.
\newblock \bibinfo{journal}{\emph{ACM transactions on graphics (TOG)}}
  \bibinfo{volume}{22}, \bibinfo{number}{3} (\bibinfo{year}{2003}),
  \bibinfo{pages}{954--961}.
\newblock


\bibitem[\protect\citeauthoryear{Kirillov, Mintun, Ravi, Mao, Rolland,
  Gustafson, Xiao, Whitehead, Berg, Lo, et~al\mbox{.}}{Kirillov
  et~al\mbox{.}}{2023}]%
        {kirillov2023segment}
\bibfield{author}{\bibinfo{person}{Alexander Kirillov}, \bibinfo{person}{Eric
  Mintun}, \bibinfo{person}{Nikhila Ravi}, \bibinfo{person}{Hanzi Mao},
  \bibinfo{person}{Chloe Rolland}, \bibinfo{person}{Laura Gustafson},
  \bibinfo{person}{Tete Xiao}, \bibinfo{person}{Spencer Whitehead},
  \bibinfo{person}{Alexander~C Berg}, \bibinfo{person}{Wan-Yen Lo},
  {et~al\mbox{.}}} \bibinfo{year}{2023}\natexlab{}.
\newblock \showarticletitle{Segment anything}. In
  \bibinfo{booktitle}{\emph{Proceedings of the IEEE/CVF international
  conference on computer vision}}. \bibinfo{pages}{4015--4026}.
\newblock


\bibitem[\protect\citeauthoryear{Koch, Matveev, Jiang, Williams, Artemov,
  Burnaev, Alexa, Zorin, and Panozzo}{Koch et~al\mbox{.}}{2019}]%
        {koch2019abc}
\bibfield{author}{\bibinfo{person}{Sebastian Koch}, \bibinfo{person}{Albert
  Matveev}, \bibinfo{person}{Zhongshi Jiang}, \bibinfo{person}{Francis
  Williams}, \bibinfo{person}{Alexey Artemov}, \bibinfo{person}{Evgeny
  Burnaev}, \bibinfo{person}{Marc Alexa}, \bibinfo{person}{Denis Zorin}, {and}
  \bibinfo{person}{Daniele Panozzo}.} \bibinfo{year}{2019}\natexlab{}.
\newblock \showarticletitle{Abc: A big cad model dataset for geometric deep
  learning}. In \bibinfo{booktitle}{\emph{Proceedings of the IEEE/CVF
  conference on computer vision and pattern recognition}}.
  \bibinfo{pages}{9601--9611}.
\newblock


\bibitem[\protect\citeauthoryear{Lavou{\'e}, Dupont, and Baskurt}{Lavou{\'e}
  et~al\mbox{.}}{2005}]%
        {lavoue2005new}
\bibfield{author}{\bibinfo{person}{Guillaume Lavou{\'e}},
  \bibinfo{person}{Florent Dupont}, {and} \bibinfo{person}{Atilla Baskurt}.}
  \bibinfo{year}{2005}\natexlab{}.
\newblock \showarticletitle{A new CAD mesh segmentation method, based on
  curvature tensor analysis}.
\newblock \bibinfo{journal}{\emph{Computer-Aided Design}} \bibinfo{volume}{37},
  \bibinfo{number}{10} (\bibinfo{year}{2005}), \bibinfo{pages}{975--987}.
\newblock


\bibitem[\protect\citeauthoryear{L\'{e}vy, Petitjean, Ray, and
  Maillot}{L\'{e}vy et~al\mbox{.}}{2002}]%
        {lscm}
\bibfield{author}{\bibinfo{person}{Bruno L\'{e}vy}, \bibinfo{person}{Sylvain
  Petitjean}, \bibinfo{person}{Nicolas Ray}, {and} \bibinfo{person}{J\'{e}rome
  Maillot}.} \bibinfo{year}{2002}\natexlab{}.
\newblock \showarticletitle{Least squares conformal maps for automatic texture
  atlas generation}.
\newblock \bibinfo{journal}{\emph{ACM Trans. Graph.}} \bibinfo{volume}{21},
  \bibinfo{number}{3} (\bibinfo{date}{July} \bibinfo{year}{2002}),
  \bibinfo{pages}{362–371}.
\newblock
\showISSN{0730-0301}
\urldef\tempurl%
\url{https://doi.org/10.1145/566654.566590}
\showDOI{\tempurl}


\bibitem[\protect\citeauthoryear{Li, Zhang, Zhang, Yang, Li, Zhong, Wang, Yuan,
  Zhang, Hwang, et~al\mbox{.}}{Li et~al\mbox{.}}{2022}]%
        {li2022grounded}
\bibfield{author}{\bibinfo{person}{Liunian~Harold Li},
  \bibinfo{person}{Pengchuan Zhang}, \bibinfo{person}{Haotian Zhang},
  \bibinfo{person}{Jianwei Yang}, \bibinfo{person}{Chunyuan Li},
  \bibinfo{person}{Yiwu Zhong}, \bibinfo{person}{Lijuan Wang},
  \bibinfo{person}{Lu Yuan}, \bibinfo{person}{Lei Zhang},
  \bibinfo{person}{Jenq-Neng Hwang}, {et~al\mbox{.}}}
  \bibinfo{year}{2022}\natexlab{}.
\newblock \showarticletitle{Grounded language-image pre-training}. In
  \bibinfo{booktitle}{\emph{Proceedings of the IEEE/CVF conference on computer
  vision and pattern recognition}}. \bibinfo{pages}{10965--10975}.
\newblock


\bibitem[\protect\citeauthoryear{Li, Kaufman, Kim, Solomon, and Sheffer}{Li
  et~al\mbox{.}}{2018}]%
        {li2018optcuts}
\bibfield{author}{\bibinfo{person}{Minchen Li}, \bibinfo{person}{Danny~M
  Kaufman}, \bibinfo{person}{Vladimir~G Kim}, \bibinfo{person}{Justin Solomon},
  {and} \bibinfo{person}{Alla Sheffer}.} \bibinfo{year}{2018}\natexlab{}.
\newblock \showarticletitle{Optcuts: Joint optimization of surface cuts and
  parameterization}.
\newblock \bibinfo{journal}{\emph{ACM transactions on graphics (TOG)}}
  \bibinfo{volume}{37}, \bibinfo{number}{6} (\bibinfo{year}{2018}),
  \bibinfo{pages}{1--13}.
\newblock


\bibitem[\protect\citeauthoryear{Liu, Zhang, Xu, Gotsman, and Gortler}{Liu
  et~al\mbox{.}}{2008}]%
        {liu2008local}
\bibfield{author}{\bibinfo{person}{Ligang Liu}, \bibinfo{person}{Lei Zhang},
  \bibinfo{person}{Yin Xu}, \bibinfo{person}{Craig Gotsman}, {and}
  \bibinfo{person}{Steven~J Gortler}.} \bibinfo{year}{2008}\natexlab{}.
\newblock \showarticletitle{A local/global approach to mesh parameterization}.
  In \bibinfo{booktitle}{\emph{Computer graphics forum}},
  Vol.~\bibinfo{volume}{27}. Wiley Online Library, \bibinfo{pages}{1495--1504}.
\newblock


\bibitem[\protect\citeauthoryear{Liu, Uy, Xiang, Su, Fidler, Sharp, and
  Gao}{Liu et~al\mbox{.}}{2025}]%
        {liu2025partfield}
\bibfield{author}{\bibinfo{person}{Minghua Liu},
  \bibinfo{person}{Mikaela~Angelina Uy}, \bibinfo{person}{Donglai Xiang},
  \bibinfo{person}{Hao Su}, \bibinfo{person}{Sanja Fidler},
  \bibinfo{person}{Nicholas Sharp}, {and} \bibinfo{person}{Jun Gao}.}
  \bibinfo{year}{2025}\natexlab{}.
\newblock \showarticletitle{PARTFIELD: Learning 3D Feature Fields for Part
  Segmentation and Beyond}.
\newblock \bibinfo{journal}{\emph{arXiv preprint arXiv:2504.11451}}
  (\bibinfo{year}{2025}).
\newblock


\bibitem[\protect\citeauthoryear{Liu, Aigerman, Kim, and Hanocka}{Liu
  et~al\mbox{.}}{2023}]%
        {liu2023wand}
\bibfield{author}{\bibinfo{person}{Richard Liu}, \bibinfo{person}{Noam
  Aigerman}, \bibinfo{person}{Vladimir~G Kim}, {and} \bibinfo{person}{Rana
  Hanocka}.} \bibinfo{year}{2023}\natexlab{}.
\newblock \showarticletitle{Da wand: Distortion-aware selection using neural
  mesh parameterization}. In \bibinfo{booktitle}{\emph{Proceedings of the
  IEEE/CVF Conference on Computer Vision and Pattern Recognition}}.
  \bibinfo{pages}{16739--16749}.
\newblock


\bibitem[\protect\citeauthoryear{Liu and Zhang}{Liu and Zhang}{2007a}]%
        {liu2007mesh}
\bibfield{author}{\bibinfo{person}{Rong Liu} {and} \bibinfo{person}{Hao
  Zhang}.} \bibinfo{year}{2007}\natexlab{a}.
\newblock \showarticletitle{Mesh segmentation via spectral embedding and
  contour analysis}. In \bibinfo{booktitle}{\emph{Computer Graphics Forum}},
  Vol.~\bibinfo{volume}{26}. Wiley Online Library, \bibinfo{pages}{385--394}.
\newblock


\bibitem[\protect\citeauthoryear{Liu and Zhang}{Liu and Zhang}{2007b}]%
        {Liu2007MeshSegmentation}
\bibfield{author}{\bibinfo{person}{Rong Liu} {and} \bibinfo{person}{Hao
  Zhang}.} \bibinfo{year}{2007}\natexlab{b}.
\newblock \showarticletitle{Mesh Segmentation via Spectral Embedding and
  Contour Analysis}.
\newblock \bibinfo{journal}{\emph{Computer Graphics Forum (Proc. Eurographics
  2007)}} \bibinfo{volume}{26}, \bibinfo{number}{3} (\bibinfo{year}{2007}),
  \bibinfo{pages}{385--394}.
\newblock
\urldef\tempurl%
\url{https://doi.org/10.1111/j.1467-8659.2007.01061.x}
\showDOI{\tempurl}


\bibitem[\protect\citeauthoryear{Lorensen and Cline}{Lorensen and
  Cline}{1998}]%
        {lorensen1998marching}
\bibfield{author}{\bibinfo{person}{William~E Lorensen} {and}
  \bibinfo{person}{Harvey~E Cline}.} \bibinfo{year}{1998}\natexlab{}.
\newblock \showarticletitle{Marching cubes: A high resolution 3D surface
  construction algorithm}.
\newblock In \bibinfo{booktitle}{\emph{Seminal graphics: pioneering efforts
  that shaped the field}}. \bibinfo{pages}{347--353}.
\newblock


\bibitem[\protect\citeauthoryear{Mo, Zhu, Chang, Yi, Tripathi, Guibas, and
  Su}{Mo et~al\mbox{.}}{2019}]%
        {mo2019partnet}
\bibfield{author}{\bibinfo{person}{Kaichun Mo}, \bibinfo{person}{Shilin Zhu},
  \bibinfo{person}{Angel~X Chang}, \bibinfo{person}{Li Yi},
  \bibinfo{person}{Subarna Tripathi}, \bibinfo{person}{Leonidas~J Guibas},
  {and} \bibinfo{person}{Hao Su}.} \bibinfo{year}{2019}\natexlab{}.
\newblock \showarticletitle{Partnet: A large-scale benchmark for fine-grained
  and hierarchical part-level 3d object understanding}. In
  \bibinfo{booktitle}{\emph{Proceedings of the IEEE/CVF conference on computer
  vision and pattern recognition}}. \bibinfo{pages}{909--918}.
\newblock


\bibitem[\protect\citeauthoryear{Morreale, Aigerman, Kim, and Mitra}{Morreale
  et~al\mbox{.}}{2021}]%
        {morreale2021neural}
\bibfield{author}{\bibinfo{person}{Luca Morreale}, \bibinfo{person}{Noam
  Aigerman}, \bibinfo{person}{Vladimir~G. Kim}, {and} \bibinfo{person}{Niloy~J.
  Mitra}.} \bibinfo{year}{2021}\natexlab{}.
\newblock \showarticletitle{Neural Surface Maps}. In
  \bibinfo{booktitle}{\emph{Conference on Computer Vision and Pattern
  Recognition}}.
\newblock


\bibitem[\protect\citeauthoryear{Mullen, Tong, Alliez, and Desbrun}{Mullen
  et~al\mbox{.}}{2008}]%
        {mullen2008spectral}
\bibfield{author}{\bibinfo{person}{Patrick Mullen}, \bibinfo{person}{Yiying
  Tong}, \bibinfo{person}{Pierre Alliez}, {and} \bibinfo{person}{Mathieu
  Desbrun}.} \bibinfo{year}{2008}\natexlab{}.
\newblock \showarticletitle{Spectral conformal parameterization}. In
  \bibinfo{booktitle}{\emph{Computer Graphics Forum}},
  Vol.~\bibinfo{volume}{27}. Wiley Online Library, \bibinfo{pages}{1487--1494}.
\newblock


\bibitem[\protect\citeauthoryear{Oh, Yuan, Wei, Shi, Xiang, Liu, and Su}{Oh
  et~al\mbox{.}}{2025}]%
        {oh2025pamo}
\bibfield{author}{\bibinfo{person}{Seonghun Oh}, \bibinfo{person}{Xiaodi Yuan},
  \bibinfo{person}{Xinyue Wei}, \bibinfo{person}{Ruoxi Shi},
  \bibinfo{person}{Fanbo Xiang}, \bibinfo{person}{Minghua Liu}, {and}
  \bibinfo{person}{Hao Su}.} \bibinfo{year}{2025}\natexlab{}.
\newblock \showarticletitle{PaMO: Parallel Mesh Optimization for
  Intersection-Free Low-Poly Modeling on the GPU}.
\newblock \bibinfo{journal}{\emph{arXiv preprint arXiv:2509.05595}}
  (\bibinfo{year}{2025}).
\newblock


\bibitem[\protect\citeauthoryear{Pietroni, Tarini, and Cignoni}{Pietroni
  et~al\mbox{.}}{2009}]%
        {pietroni2009almost}
\bibfield{author}{\bibinfo{person}{Nico Pietroni}, \bibinfo{person}{Marco
  Tarini}, {and} \bibinfo{person}{Paolo Cignoni}.}
  \bibinfo{year}{2009}\natexlab{}.
\newblock \showarticletitle{Almost isometric mesh parameterization through
  abstract domains}.
\newblock \bibinfo{journal}{\emph{IEEE Transactions on Visualization and
  Computer Graphics}} \bibinfo{volume}{16}, \bibinfo{number}{4}
  (\bibinfo{year}{2009}), \bibinfo{pages}{621--635}.
\newblock


\bibitem[\protect\citeauthoryear{Poranne, Tarini, Huber, Panozzo, and
  Sorkine-Hornung}{Poranne et~al\mbox{.}}{2017}]%
        {autoCuts}
\bibfield{author}{\bibinfo{person}{Roi Poranne}, \bibinfo{person}{Marco
  Tarini}, \bibinfo{person}{Sandro Huber}, \bibinfo{person}{Daniele Panozzo},
  {and} \bibinfo{person}{Olga Sorkine-Hornung}.}
  \bibinfo{year}{2017}\natexlab{}.
\newblock \showarticletitle{{AutoCuts: Simultaneous Distortion and Cut
  Optimization for UV Mapping}}.
\newblock \bibinfo{journal}{\emph{ACM Transactions on Graphics}}
  (\bibinfo{year}{2017}).
\newblock


\bibitem[\protect\citeauthoryear{Pothen, Simon, and Liou}{Pothen
  et~al\mbox{.}}{1990}]%
        {pothen1990partitioning}
\bibfield{author}{\bibinfo{person}{Alex Pothen}, \bibinfo{person}{Horst~D
  Simon}, {and} \bibinfo{person}{Kang-Pu Liou}.}
  \bibinfo{year}{1990}\natexlab{}.
\newblock \showarticletitle{Partitioning sparse matrices with eigenvectors of
  graphs}.
\newblock \bibinfo{journal}{\emph{SIAM journal on matrix analysis and
  applications}} \bibinfo{volume}{11}, \bibinfo{number}{3}
  (\bibinfo{year}{1990}), \bibinfo{pages}{430--452}.
\newblock


\bibitem[\protect\citeauthoryear{Pulla, Razdan, and Farin}{Pulla
  et~al\mbox{.}}{2001}]%
        {pulla2001improved}
\bibfield{author}{\bibinfo{person}{Sandeep Pulla}, \bibinfo{person}{Anshuman
  Razdan}, {and} \bibinfo{person}{Gerald Farin}.}
  \bibinfo{year}{2001}\natexlab{}.
\newblock \showarticletitle{Improved curvature estimation for watershed
  segmentation of 3-dimensional meshes}.
\newblock \bibinfo{journal}{\emph{IEEE Transactions on Visualization and
  Computer Graphics}} \bibinfo{volume}{5}, \bibinfo{number}{4}
  (\bibinfo{year}{2001}), \bibinfo{pages}{308--321}.
\newblock


\bibitem[\protect\citeauthoryear{Qian, Li, Peng, Mai, Hammoud, Elhoseiny, and
  Ghanem}{Qian et~al\mbox{.}}{2022}]%
        {qian2022pointnext}
\bibfield{author}{\bibinfo{person}{Guocheng Qian}, \bibinfo{person}{Yuchen Li},
  \bibinfo{person}{Houwen Peng}, \bibinfo{person}{Jinjie Mai},
  \bibinfo{person}{Hasan Hammoud}, \bibinfo{person}{Mohamed Elhoseiny}, {and}
  \bibinfo{person}{Bernard Ghanem}.} \bibinfo{year}{2022}\natexlab{}.
\newblock \showarticletitle{Pointnext: Revisiting pointnet++ with improved
  training and scaling strategies}.
\newblock \bibinfo{journal}{\emph{Advances in neural information processing
  systems}}  \bibinfo{volume}{35} (\bibinfo{year}{2022}),
  \bibinfo{pages}{23192--23204}.
\newblock


\bibitem[\protect\citeauthoryear{Rabinovich, Poranne, Panozzo, and
  Sorkine-Hornung}{Rabinovich et~al\mbox{.}}{2017}]%
        {rabinovich2017scalable}
\bibfield{author}{\bibinfo{person}{Michael Rabinovich}, \bibinfo{person}{Roi
  Poranne}, \bibinfo{person}{Daniele Panozzo}, {and} \bibinfo{person}{Olga
  Sorkine-Hornung}.} \bibinfo{year}{2017}\natexlab{}.
\newblock \showarticletitle{Scalable locally injective mappings}.
\newblock \bibinfo{journal}{\emph{ACM Transactions on Graphics (TOG)}}
  \bibinfo{volume}{36}, \bibinfo{number}{4} (\bibinfo{year}{2017}),
  \bibinfo{pages}{1}.
\newblock


\bibitem[\protect\citeauthoryear{Ray and L{\'e}vy}{Ray and L{\'e}vy}{2003}]%
        {Ray2003HierarchicalLS}
\bibfield{author}{\bibinfo{person}{Nicolas Ray} {and} \bibinfo{person}{Bruno
  L{\'e}vy}.} \bibinfo{year}{2003}\natexlab{}.
\newblock \showarticletitle{Hierarchical least squares conformal map}.
\newblock \bibinfo{journal}{\emph{11th Pacific Conference onComputer Graphics
  and Applications, 2003. Proceedings.}} (\bibinfo{year}{2003}),
  \bibinfo{pages}{263--270}.
\newblock
\urldef\tempurl%
\url{https://api.semanticscholar.org/CorpusID:1121017}
\showURL{%
\tempurl}


\bibitem[\protect\citeauthoryear{Roy}{Roy}{2023}]%
        {roy2023neural}
\bibfield{author}{\bibinfo{person}{Bruno Roy}.}
  \bibinfo{year}{2023}\natexlab{}.
\newblock \showarticletitle{Neural ShDF: Reviving an Efficient and Consistent
  Mesh Segmentation Method}.
\newblock \bibinfo{journal}{\emph{arXiv preprint arXiv:2306.11737}}
  (\bibinfo{year}{2023}).
\newblock


\bibitem[\protect\citeauthoryear{Sander, Gortler, Snyder, and Hoppe}{Sander
  et~al\mbox{.}}{2002}]%
        {sander2002signal}
\bibfield{author}{\bibinfo{person}{Pedro~V Sander}, \bibinfo{person}{Steven
  Gortler}, \bibinfo{person}{John Snyder}, {and} \bibinfo{person}{Hugues
  Hoppe}.} \bibinfo{year}{2002}\natexlab{}.
\newblock \showarticletitle{Signal-specialized parameterization}.
\newblock  (\bibinfo{year}{2002}).
\newblock


\bibitem[\protect\citeauthoryear{Sander, Snyder, Gortler, and Hoppe}{Sander
  et~al\mbox{.}}{2001}]%
        {sander2001texture}
\bibfield{author}{\bibinfo{person}{Pedro~V Sander}, \bibinfo{person}{John
  Snyder}, \bibinfo{person}{Steven~J Gortler}, {and} \bibinfo{person}{Hugues
  Hoppe}.} \bibinfo{year}{2001}\natexlab{}.
\newblock \showarticletitle{Texture mapping progressive meshes}. In
  \bibinfo{booktitle}{\emph{Proceedings of the 28th annual conference on
  Computer graphics and interactive techniques}}. \bibinfo{pages}{409--416}.
\newblock


\bibitem[\protect\citeauthoryear{Sander, Wood, Gortler, Snyder, and
  Hoppe}{Sander et~al\mbox{.}}{2003}]%
        {Sander2003MultiChartGI}
\bibfield{author}{\bibinfo{person}{Pedro~V. Sander},
  \bibinfo{person}{Zo{\"e}~J. Wood}, \bibinfo{person}{Steven~J. Gortler},
  \bibinfo{person}{John~M. Snyder}, {and} \bibinfo{person}{Hugues Hoppe}.}
  \bibinfo{year}{2003}\natexlab{}.
\newblock \showarticletitle{Multi-Chart Geometry Images}. In
  \bibinfo{booktitle}{\emph{Eurographics Symposium on Geometry Processing}}.
\newblock
\urldef\tempurl%
\url{https://api.semanticscholar.org/CorpusID:4893489}
\showURL{%
\tempurl}


\bibitem[\protect\citeauthoryear{Sawhney and Crane}{Sawhney and Crane}{2017}]%
        {sawhney2017boundary}
\bibfield{author}{\bibinfo{person}{Rohan Sawhney} {and} \bibinfo{person}{Keenan
  Crane}.} \bibinfo{year}{2017}\natexlab{}.
\newblock \showarticletitle{Boundary first flattening}.
\newblock \bibinfo{journal}{\emph{ACM Transactions on Graphics (ToG)}}
  \bibinfo{volume}{37}, \bibinfo{number}{1} (\bibinfo{year}{2017}),
  \bibinfo{pages}{1--14}.
\newblock


\bibitem[\protect\citeauthoryear{Sch{\"u}ller, Kavan, Panozzo, and
  Sorkine-Hornung}{Sch{\"u}ller et~al\mbox{.}}{2013}]%
        {schuller2013locally}
\bibfield{author}{\bibinfo{person}{Christian Sch{\"u}ller},
  \bibinfo{person}{Ladislav Kavan}, \bibinfo{person}{Daniele Panozzo}, {and}
  \bibinfo{person}{Olga Sorkine-Hornung}.} \bibinfo{year}{2013}\natexlab{}.
\newblock \showarticletitle{Locally injective mappings}. In
  \bibinfo{booktitle}{\emph{Computer Graphics Forum}},
  Vol.~\bibinfo{volume}{32}. Wiley Online Library, \bibinfo{pages}{125--135}.
\newblock


\bibitem[\protect\citeauthoryear{Sheffer and De~Sturler}{Sheffer and
  De~Sturler}{2000}]%
        {sheffer2000surface}
\bibfield{author}{\bibinfo{person}{Alla Sheffer} {and} \bibinfo{person}{Eric
  De~Sturler}.} \bibinfo{year}{2000}\natexlab{}.
\newblock \bibinfo{title}{Surface Parameterization for Meshing by Triangulation
  Flattenin}.
\newblock
\newblock


\bibitem[\protect\citeauthoryear{Sheffer and de~Sturler}{Sheffer and
  de~Sturler}{2001}]%
        {sheffer2001parameterization}
\bibfield{author}{\bibinfo{person}{Alla Sheffer} {and} \bibinfo{person}{Eric de
  Sturler}.} \bibinfo{year}{2001}\natexlab{}.
\newblock \showarticletitle{Parameterization of faceted surfaces for meshing
  using angle-based flattening}.
\newblock \bibinfo{journal}{\emph{Engineering with computers}}
  \bibinfo{volume}{17} (\bibinfo{year}{2001}), \bibinfo{pages}{326--337}.
\newblock


\bibitem[\protect\citeauthoryear{Sheffer, L{\'e}vy, Mogilnitsky, and
  Bogomyakov}{Sheffer et~al\mbox{.}}{2005}]%
        {sheffer2005abf++}
\bibfield{author}{\bibinfo{person}{Alla Sheffer}, \bibinfo{person}{Bruno
  L{\'e}vy}, \bibinfo{person}{Maxim Mogilnitsky}, {and}
  \bibinfo{person}{Alexander Bogomyakov}.} \bibinfo{year}{2005}\natexlab{}.
\newblock \showarticletitle{ABF++: fast and robust angle based flattening}.
\newblock \bibinfo{journal}{\emph{ACM Transactions on Graphics (TOG)}}
  \bibinfo{volume}{24}, \bibinfo{number}{2} (\bibinfo{year}{2005}),
  \bibinfo{pages}{311--330}.
\newblock


\bibitem[\protect\citeauthoryear{Sorkine and Alexa}{Sorkine and Alexa}{2007}]%
        {sorkine2007rigid}
\bibfield{author}{\bibinfo{person}{Olga Sorkine} {and} \bibinfo{person}{Marc
  Alexa}.} \bibinfo{year}{2007}\natexlab{}.
\newblock \showarticletitle{As-rigid-as-possible surface modeling}. In
  \bibinfo{booktitle}{\emph{Symposium on Geometry processing}},
  Vol.~\bibinfo{volume}{4}. Citeseer, \bibinfo{pages}{109--116}.
\newblock


\bibitem[\protect\citeauthoryear{Sorkine, Cohen-Or, Goldenthal, and
  Lischinski}{Sorkine et~al\mbox{.}}{2002}]%
        {sorkine2002bounded}
\bibfield{author}{\bibinfo{person}{Olga Sorkine}, \bibinfo{person}{Daniel
  Cohen-Or}, \bibinfo{person}{Rony Goldenthal}, {and} \bibinfo{person}{Dani
  Lischinski}.} \bibinfo{year}{2002}\natexlab{}.
\newblock \showarticletitle{Bounded-distortion piecewise mesh
  parameterization}. In \bibinfo{booktitle}{\emph{IEEE Visualization, 2002. VIS
  2002.}} IEEE, \bibinfo{pages}{355--362}.
\newblock


\bibitem[\protect\citeauthoryear{Srinivasan, Garbin, Verbin, Barron, and
  Mildenhall}{Srinivasan et~al\mbox{.}}{2024}]%
        {srinivasan2024nuvo}
\bibfield{author}{\bibinfo{person}{Pratul~P Srinivasan},
  \bibinfo{person}{Stephan~J Garbin}, \bibinfo{person}{Dor Verbin},
  \bibinfo{person}{Jonathan~T Barron}, {and} \bibinfo{person}{Ben Mildenhall}.}
  \bibinfo{year}{2024}\natexlab{}.
\newblock \showarticletitle{Nuvo: Neural uv mapping for unruly 3d
  representations}. In \bibinfo{booktitle}{\emph{European Conference on
  Computer Vision}}. Springer, \bibinfo{pages}{18--34}.
\newblock


\bibitem[\protect\citeauthoryear{Su, Cui, Qian, Lei, Zhang, Zhang, and Gu}{Su
  et~al\mbox{.}}{2016}]%
        {su2016area}
\bibfield{author}{\bibinfo{person}{Kehua Su}, \bibinfo{person}{Li Cui},
  \bibinfo{person}{Kun Qian}, \bibinfo{person}{Na Lei}, \bibinfo{person}{Junwei
  Zhang}, \bibinfo{person}{Min Zhang}, {and} \bibinfo{person}{Xianfeng~David
  Gu}.} \bibinfo{year}{2016}\natexlab{}.
\newblock \showarticletitle{Area-preserving mesh parameterization for
  poly-annulus surfaces based on optimal mass transportation}.
\newblock \bibinfo{journal}{\emph{Computer Aided Geometric Design}}
  \bibinfo{volume}{46} (\bibinfo{year}{2016}), \bibinfo{pages}{76--91}.
\newblock


\bibitem[\protect\citeauthoryear{Takahashi, Wu, Saw, Lin, and Yen}{Takahashi
  et~al\mbox{.}}{2011}]%
        {takahashi2011optimized}
\bibfield{author}{\bibinfo{person}{Shigeo Takahashi},
  \bibinfo{person}{Hsiang-Yun Wu}, \bibinfo{person}{Seow~Hui Saw},
  \bibinfo{person}{Chun-Cheng Lin}, {and} \bibinfo{person}{Hsu-Chun Yen}.}
  \bibinfo{year}{2011}\natexlab{}.
\newblock \showarticletitle{Optimized topological surgery for unfolding 3d
  meshes}. In \bibinfo{booktitle}{\emph{Computer graphics forum}},
  Vol.~\bibinfo{volume}{30}. Wiley Online Library, \bibinfo{pages}{2077--2086}.
\newblock


\bibitem[\protect\citeauthoryear{Taubin}{Taubin}{1995}]%
        {taubin1995signal}
\bibfield{author}{\bibinfo{person}{Gabriel Taubin}.}
  \bibinfo{year}{1995}\natexlab{}.
\newblock \showarticletitle{A signal processing approach to fair surface
  design}. In \bibinfo{booktitle}{\emph{Proceedings of the 22nd annual
  conference on Computer graphics and interactive techniques}}.
  \bibinfo{pages}{351--358}.
\newblock


\bibitem[\protect\citeauthoryear{Tutte}{Tutte}{1963}]%
        {tutte1963draw}
\bibfield{author}{\bibinfo{person}{William~Thomas Tutte}.}
  \bibinfo{year}{1963}\natexlab{}.
\newblock \showarticletitle{How to draw a graph}.
\newblock \bibinfo{journal}{\emph{Proceedings of the London Mathematical
  Society}} \bibinfo{volume}{3}, \bibinfo{number}{1} (\bibinfo{year}{1963}),
  \bibinfo{pages}{743--767}.
\newblock


\bibitem[\protect\citeauthoryear{Vu, Kim, Luu, Nguyen, and Yoo}{Vu
  et~al\mbox{.}}{2022}]%
        {vu2022softgroup}
\bibfield{author}{\bibinfo{person}{Thang Vu}, \bibinfo{person}{Kookhoi Kim},
  \bibinfo{person}{Tung~M Luu}, \bibinfo{person}{Thanh Nguyen}, {and}
  \bibinfo{person}{Chang~D Yoo}.} \bibinfo{year}{2022}\natexlab{}.
\newblock \showarticletitle{Softgroup for 3d instance segmentation on point
  clouds}. In \bibinfo{booktitle}{\emph{Proceedings of the IEEE/CVF conference
  on computer vision and pattern recognition}}. \bibinfo{pages}{2708--2717}.
\newblock


\bibitem[\protect\citeauthoryear{Xiang, Xu, Hasan, Hold-Geoffroy, Sunkavalli,
  and Su}{Xiang et~al\mbox{.}}{2021}]%
        {xiang2021neutex}
\bibfield{author}{\bibinfo{person}{Fanbo Xiang}, \bibinfo{person}{Zexiang Xu},
  \bibinfo{person}{Milos Hasan}, \bibinfo{person}{Yannick Hold-Geoffroy},
  \bibinfo{person}{Kalyan Sunkavalli}, {and} \bibinfo{person}{Hao Su}.}
  \bibinfo{year}{2021}\natexlab{}.
\newblock \showarticletitle{Neutex: Neural texture mapping for volumetric
  neural rendering}. In \bibinfo{booktitle}{\emph{Proceedings of the IEEE/CVF
  Conference on Computer Vision and Pattern Recognition}}.
  \bibinfo{pages}{7119--7128}.
\newblock


\bibitem[\protect\citeauthoryear{Xiang, Lv, Xu, Deng, Wang, Zhang, Chen, Tong,
  and Yang}{Xiang et~al\mbox{.}}{2024}]%
        {xiang2024structured}
\bibfield{author}{\bibinfo{person}{Jianfeng Xiang}, \bibinfo{person}{Zelong
  Lv}, \bibinfo{person}{Sicheng Xu}, \bibinfo{person}{Yu Deng},
  \bibinfo{person}{Ruicheng Wang}, \bibinfo{person}{Bowen Zhang},
  \bibinfo{person}{Dong Chen}, \bibinfo{person}{Xin Tong}, {and}
  \bibinfo{person}{Jiaolong Yang}.} \bibinfo{year}{2024}\natexlab{}.
\newblock \showarticletitle{Structured 3d latents for scalable and versatile 3d
  generation}.
\newblock \bibinfo{journal}{\emph{arXiv preprint arXiv:2412.01506}}
  (\bibinfo{year}{2024}).
\newblock


\bibitem[\protect\citeauthoryear{Xu, Yin, Qiu, Liu, Tong, and Han}{Xu
  et~al\mbox{.}}{2023}]%
        {xu2023sampro3d}
\bibfield{author}{\bibinfo{person}{Mutian Xu}, \bibinfo{person}{Xingyilang
  Yin}, \bibinfo{person}{Lingteng Qiu}, \bibinfo{person}{Yang Liu},
  \bibinfo{person}{Xin Tong}, {and} \bibinfo{person}{Xiaoguang Han}.}
  \bibinfo{year}{2023}\natexlab{}.
\newblock \showarticletitle{Sampro3d: Locating sam prompts in 3d for zero-shot
  scene segmentation}.
\newblock \bibinfo{journal}{\emph{arXiv preprint arXiv:2311.17707}}
  (\bibinfo{year}{2023}).
\newblock


\bibitem[\protect\citeauthoryear{Xu, Chen, Zhao, Wang, Zhou, and Lu}{Xu
  et~al\mbox{.}}{2024}]%
        {xu2024embodiedsam}
\bibfield{author}{\bibinfo{person}{Xiuwei Xu}, \bibinfo{person}{Huangxing
  Chen}, \bibinfo{person}{Linqing Zhao}, \bibinfo{person}{Ziwei Wang},
  \bibinfo{person}{Jie Zhou}, {and} \bibinfo{person}{Jiwen Lu}.}
  \bibinfo{year}{2024}\natexlab{}.
\newblock \showarticletitle{Embodiedsam: Online segment any 3d thing in real
  time}.
\newblock \bibinfo{journal}{\emph{arXiv preprint arXiv:2408.11811}}
  (\bibinfo{year}{2024}).
\newblock


\bibitem[\protect\citeauthoryear{Yamauchi, Gumhold, Zayer, and Seidel}{Yamauchi
  et~al\mbox{.}}{2005}]%
        {yamauchi2005mesh}
\bibfield{author}{\bibinfo{person}{Hitoshi Yamauchi}, \bibinfo{person}{Stefan
  Gumhold}, \bibinfo{person}{Rhaleb Zayer}, {and} \bibinfo{person}{Hans-Peter
  Seidel}.} \bibinfo{year}{2005}\natexlab{}.
\newblock \showarticletitle{Mesh segmentation driven by Gaussian curvature}.
\newblock \bibinfo{journal}{\emph{The Visual Computer}}  \bibinfo{volume}{21}
  (\bibinfo{year}{2005}), \bibinfo{pages}{659--668}.
\newblock


\bibitem[\protect\citeauthoryear{Yan, Yang, Shi, and Zhang}{Yan
  et~al\mbox{.}}{2005}]%
        {yan2005mesh}
\bibfield{author}{\bibinfo{person}{Jingqi Yan}, \bibinfo{person}{Xin Yang},
  \bibinfo{person}{Pengfei Shi}, {and} \bibinfo{person}{David Zhang}.}
  \bibinfo{year}{2005}\natexlab{}.
\newblock \showarticletitle{Mesh parameterization by minimizing the synthesized
  distortion metric with the coefficient-optimizing algorithm}.
\newblock \bibinfo{journal}{\emph{IEEE transactions on visualization and
  computer graphics}} \bibinfo{volume}{12}, \bibinfo{number}{1}
  (\bibinfo{year}{2005}), \bibinfo{pages}{83--92}.
\newblock


\bibitem[\protect\citeauthoryear{Yang, Huang, Guo, Lu, Wu, Lam, Cao, and
  Liu}{Yang et~al\mbox{.}}{2024}]%
        {yang2024sampart3d}
\bibfield{author}{\bibinfo{person}{Yunhan Yang}, \bibinfo{person}{Yukun Huang},
  \bibinfo{person}{Yuan-Chen Guo}, \bibinfo{person}{Liangjun Lu},
  \bibinfo{person}{Xiaoyang Wu}, \bibinfo{person}{Edmund~Y Lam},
  \bibinfo{person}{Yan-Pei Cao}, {and} \bibinfo{person}{Xihui Liu}.}
  \bibinfo{year}{2024}\natexlab{}.
\newblock \showarticletitle{Sampart3d: Segment any part in 3d objects}.
\newblock \bibinfo{journal}{\emph{arXiv preprint arXiv:2411.07184}}
  (\bibinfo{year}{2024}).
\newblock


\bibitem[\protect\citeauthoryear{Yang, Wu, He, Zhao, and Liu}{Yang
  et~al\mbox{.}}{2023}]%
        {yang2023sam3d}
\bibfield{author}{\bibinfo{person}{Yunhan Yang}, \bibinfo{person}{Xiaoyang Wu},
  \bibinfo{person}{Tong He}, \bibinfo{person}{Hengshuang Zhao}, {and}
  \bibinfo{person}{Xihui Liu}.} \bibinfo{year}{2023}\natexlab{}.
\newblock \showarticletitle{Sam3d: Segment anything in 3d scenes}.
\newblock \bibinfo{journal}{\emph{arXiv preprint arXiv:2306.03908}}
  (\bibinfo{year}{2023}).
\newblock


\bibitem[\protect\citeauthoryear{Yi, Kim, Ceylan, Shen, Yan, Su, Lu, Huang,
  Sheffer, and Guibas}{Yi et~al\mbox{.}}{2016}]%
        {yi2016scalable}
\bibfield{author}{\bibinfo{person}{Li Yi}, \bibinfo{person}{Vladimir~G Kim},
  \bibinfo{person}{Duygu Ceylan}, \bibinfo{person}{I-Chao Shen},
  \bibinfo{person}{Mengyan Yan}, \bibinfo{person}{Hao Su},
  \bibinfo{person}{Cewu Lu}, \bibinfo{person}{Qixing Huang},
  \bibinfo{person}{Alla Sheffer}, {and} \bibinfo{person}{Leonidas Guibas}.}
  \bibinfo{year}{2016}\natexlab{}.
\newblock \showarticletitle{A scalable active framework for region annotation
  in 3d shape collections}.
\newblock \bibinfo{journal}{\emph{ACM Transactions on Graphics (ToG)}}
  \bibinfo{volume}{35}, \bibinfo{number}{6} (\bibinfo{year}{2016}),
  \bibinfo{pages}{1--12}.
\newblock


\bibitem[\protect\citeauthoryear{Yin, Liu, Xiao, Cohen-Or, Huang, and Chen}{Yin
  et~al\mbox{.}}{2024}]%
        {yin2024sai3d}
\bibfield{author}{\bibinfo{person}{Yingda Yin}, \bibinfo{person}{Yuzheng Liu},
  \bibinfo{person}{Yang Xiao}, \bibinfo{person}{Daniel Cohen-Or},
  \bibinfo{person}{Jingwei Huang}, {and} \bibinfo{person}{Baoquan Chen}.}
  \bibinfo{year}{2024}\natexlab{}.
\newblock \showarticletitle{Sai3d: Segment any instance in 3d scenes}. In
  \bibinfo{booktitle}{\emph{Proceedings of the IEEE/CVF Conference on Computer
  Vision and Pattern Recognition}}. \bibinfo{pages}{3292--3302}.
\newblock


\bibitem[\protect\citeauthoryear{Young}{Young}{2019}]%
        {xatlas}
\bibfield{author}{\bibinfo{person}{Jonathan Young}.}
  \bibinfo{year}{2019}\natexlab{}.
\newblock \bibinfo{title}{xatlas}.
\newblock \bibinfo{howpublished}{\url{https://github.com/jpcy/xatlas}}.
\newblock
\newblock
\shownote{Accessed: 2025-09-26.}


\bibitem[\protect\citeauthoryear{Yueh, Lin, Wu, and Yau}{Yueh
  et~al\mbox{.}}{2019}]%
        {yueh2019novel}
\bibfield{author}{\bibinfo{person}{Mei-Heng Yueh}, \bibinfo{person}{Wen-Wei
  Lin}, \bibinfo{person}{Chin-Tien Wu}, {and} \bibinfo{person}{Shing-Tung
  Yau}.} \bibinfo{year}{2019}\natexlab{}.
\newblock \showarticletitle{A novel stretch energy minimization algorithm for
  equiareal parameterizations}.
\newblock \bibinfo{journal}{\emph{Journal of Scientific Computing}}
  \bibinfo{volume}{78} (\bibinfo{year}{2019}), \bibinfo{pages}{1353--1386}.
\newblock


\bibitem[\protect\citeauthoryear{Zayer, L{\'e}vy, and Seidel}{Zayer
  et~al\mbox{.}}{2007}]%
        {zayer2007linear}
\bibfield{author}{\bibinfo{person}{Rhaleb Zayer}, \bibinfo{person}{Bruno
  L{\'e}vy}, {and} \bibinfo{person}{Hans-Peter Seidel}.}
  \bibinfo{year}{2007}\natexlab{}.
\newblock \showarticletitle{Linear angle based parameterization}. In
  \bibinfo{booktitle}{\emph{Fifth Eurographics Symposium on Geometry
  Processing-SGP 2007}}. Eurographics Association, \bibinfo{pages}{135--141}.
\newblock


\bibitem[\protect\citeauthoryear{Zhang, Mischaikow, and Turk}{Zhang
  et~al\mbox{.}}{2005}]%
        {zhang2005feature}
\bibfield{author}{\bibinfo{person}{Eugene Zhang}, \bibinfo{person}{Konstantin
  Mischaikow}, {and} \bibinfo{person}{Greg Turk}.}
  \bibinfo{year}{2005}\natexlab{}.
\newblock \showarticletitle{Feature-based surface parameterization and texture
  mapping}.
\newblock \bibinfo{journal}{\emph{ACM Transactions on Graphics (TOG)}}
  \bibinfo{volume}{24}, \bibinfo{number}{1} (\bibinfo{year}{2005}),
  \bibinfo{pages}{1--27}.
\newblock


\bibitem[\protect\citeauthoryear{Zhang, Hou, Wang, and He}{Zhang
  et~al\mbox{.}}{2024}]%
        {zhang2024flatten}
\bibfield{author}{\bibinfo{person}{Qijian Zhang}, \bibinfo{person}{Junhui Hou},
  \bibinfo{person}{Wenping Wang}, {and} \bibinfo{person}{Ying He}.}
  \bibinfo{year}{2024}\natexlab{}.
\newblock \showarticletitle{Flatten Anything: Unsupervised Neural Surface
  Parameterization}. In \bibinfo{booktitle}{\emph{Proc. NeurIPS}}.
\newblock


\bibitem[\protect\citeauthoryear{Zhao, Zhang, Hou, Xia, Wang, and He}{Zhao
  et~al\mbox{.}}{2025}]%
        {zhao2025flexpara}
\bibfield{author}{\bibinfo{person}{Yuming Zhao}, \bibinfo{person}{Qijian
  Zhang}, \bibinfo{person}{Junhui Hou}, \bibinfo{person}{Jiazhi Xia},
  \bibinfo{person}{Wenping Wang}, {and} \bibinfo{person}{Ying He}.}
  \bibinfo{year}{2025}\natexlab{}.
\newblock \showarticletitle{FlexPara: Flexible Neural Surface
  Parameterization}.
\newblock \bibinfo{journal}{\emph{arXiv preprint arXiv:2504.19210}}
  (\bibinfo{year}{2025}).
\newblock


\bibitem[\protect\citeauthoryear{Zhou, Synder, Guo, and Shum}{Zhou
  et~al\mbox{.}}{2004}]%
        {zhou2004iso}
\bibfield{author}{\bibinfo{person}{Kun Zhou}, \bibinfo{person}{John Synder},
  \bibinfo{person}{Baining Guo}, {and} \bibinfo{person}{Heung-Yeung Shum}.}
  \bibinfo{year}{2004}\natexlab{}.
\newblock \showarticletitle{Iso-charts: stretch-driven mesh parameterization
  using spectral analysis}. In \bibinfo{booktitle}{\emph{Proceedings of the
  2004 Eurographics/ACM SIGGRAPH symposium on Geometry processing}}.
  \bibinfo{pages}{45--54}.
\newblock


\bibitem[\protect\citeauthoryear{Zhou, Park, and Koltun}{Zhou
  et~al\mbox{.}}{2018a}]%
        {zhou2018open3d}
\bibfield{author}{\bibinfo{person}{Qian-Yi Zhou}, \bibinfo{person}{Jaesik
  Park}, {and} \bibinfo{person}{Vladlen Koltun}.}
  \bibinfo{year}{2018}\natexlab{a}.
\newblock \showarticletitle{Open3D: A modern library for 3D data processing}.
\newblock \bibinfo{journal}{\emph{arXiv preprint arXiv:1801.09847}}
  (\bibinfo{year}{2018}).
\newblock


\bibitem[\protect\citeauthoryear{Zhou, Park, and Koltun}{Zhou
  et~al\mbox{.}}{2018b}]%
        {Zhou2018}
\bibfield{author}{\bibinfo{person}{Qian-Yi Zhou}, \bibinfo{person}{Jaesik
  Park}, {and} \bibinfo{person}{Vladlen Koltun}.}
  \bibinfo{year}{2018}\natexlab{b}.
\newblock \showarticletitle{{Open3D}: {A} Modern Library for {3D} Data
  Processing}.
\newblock \bibinfo{journal}{\emph{arXiv:1801.09847}} (\bibinfo{year}{2018}).
\newblock


\bibitem[\protect\citeauthoryear{Łukasz Czyż}{Łukasz Czyż}{2025}]%
        {uvpackmaster3}
\bibfield{author}{\bibinfo{person}{3~Coords~Computing Łukasz Czyż}.}
  \bibinfo{year}{2025}\natexlab{}.
\newblock \bibinfo{title}{UVPackmaster 3: GPU Accelerated UV Packing Engine}.
\newblock \bibinfo{howpublished}{\url{https://uvpackmaster.com/}}.
\newblock
\newblock
\shownote{Accessed: 2025-05-21.}


\end{thebibliography}
\end{document}